\documentclass[review, compsoc]{IEEEtran}
\usepackage{amsfonts}
\usepackage{amsmath}
\usepackage{amsmath,epsfig,amsmath}
\usepackage{amsfonts,amssymb,amsthm}
\usepackage{epsfig}
\usepackage{float}
\usepackage{graphicx}
\usepackage{color,hyperref}
\usepackage{multirow}
\usepackage{tipa}
\usepackage{picins}
\usepackage{color}
\usepackage{booktabs}
\usepackage{threeparttable}
\usepackage{makecell}

\begin{document}

\title{Behavioral Intention Prediction \\in Driving Scenes: A Survey}

\author{Jianwu Fang, Fan Wang, Jianru Xue, and Tat-Seng Chua
\thanks{J. Fang and J. Xue are with the Institute of Artificial Intelligence and Robotics, Xi'an Jiaotong University, Xi'an, China
        {(fangjianwu@ieee.org, jrxue@mail.xjtu.edu.cn)}.}%
\thanks{F. Wang is with the College of Transportation Engineering, Chang'an University, Xi'an, China.}%   
        \thanks{T. Chua is with the Sea-NExT++ Joint Research Centre of the School of Computing, National University of Singapore, Singapore
        {(dcscts@nus.edu.sg).}}%   
}

\markboth{IEEE Latex}%
{Shell \MakeLowercase{\textit{et al.}}: Bare Demo of IEEEtran.cls for Computer Society Journals}

\IEEEtitleabstractindextext{%
\begin{abstract}
In the driving scene, the road agents usually conduct frequent interaction and intention understanding of the surroundings. Ego-agent (each road agent itself) predicts what behavior will be engaged by other road users all the time and expects a shared and consistent understanding for safe movement. Behavioral Intention Prediction (BIP) simulates such a human consideration process and fulfills the early prediction of specific behaviors. Similar to other prediction tasks, such as trajectory prediction, data-driven deep learning methods have taken the primary pipeline in research. The rapid development of BIP inevitably leads to new issues and challenges. To catalyze future research, this work provides a comprehensive review of BIP from the available datasets, key factors and challenges, pedestrian-centric and vehicle-centric BIP approaches, and BIP-aware applications. Based on the investigation, data-driven deep learning approaches have become the primary pipelines. The behavioral intention types are still monotonous in most current datasets and methods (\emph{e.g.}, Crossing (C) and Not Crossing (NC) for pedestrians and Lane Changing (LC) for vehicles) in this field. In addition, for the safe-critical scenarios (\emph{e.g.}, near-crashing situations), current research is limited. Through this investigation, we identify open issues in behavioral intention prediction and suggest possible insights for future research.
\end{abstract}

\begin{IEEEkeywords}
Behavioral intention prediction, challenges, promising approaches, road agents, benchmarks
\end{IEEEkeywords}}

\maketitle

\IEEEdisplaynontitleabstractindextext

% \fi
\IEEEpeerreviewmaketitle

\IEEEraisesectionheading{
\section{Introduction}
\label{section1}}
\IEEEPARstart{T}{he} driving scene is highly socialized which necessitates the effective and precise understanding of the intentions of surrounding road agents. For safe driving, the decision-making is influenced by any actions made by target road pedestrians, vehicles, cyclists, \emph{etc.} The \emph{behavioral intention} in driving scenes links the anticipated actions or behaviors of the road agents, such as ``\emph{crossing the street}" for pedestrians/cyclists and ``\emph{changing lanes}" for vehicles. It reveals the deliberate tendency of road agents to take specific actions or achieve a specific goal, which is usually understood as the internal reason for presenting specific behaviors \cite{velleman1991intention,malle1997folk}. 

 \begin{figure}[!t]
  \centering
 \includegraphics[width=\hsize]{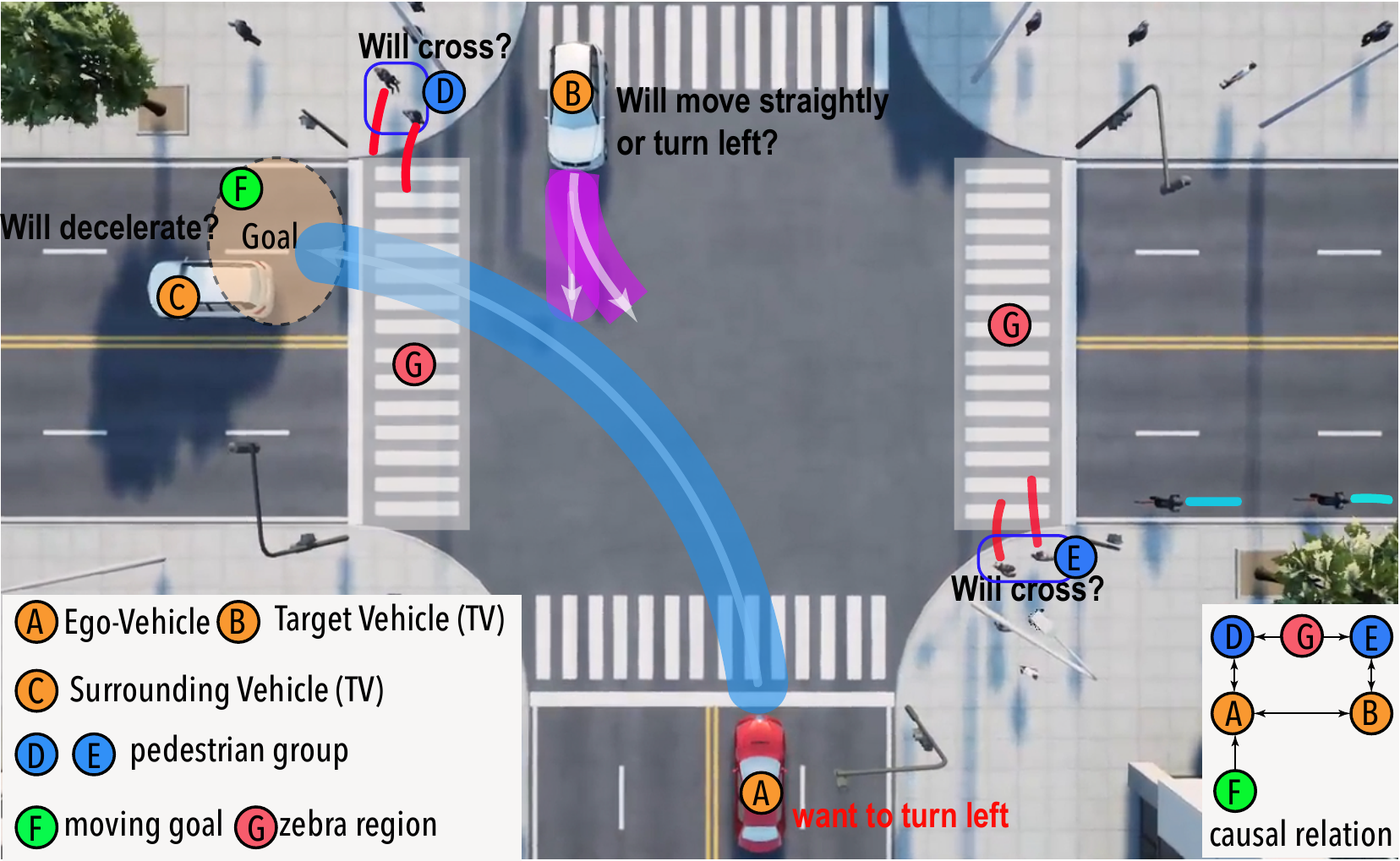}
  \caption{An intersection scenario, where the Ego Vehicle ($A$) wants to turn left and arrives at the goal region ($F$). It needs to estimate whether the Target Vehicle ($B$) will move straight or turn left, and the crossing intention of the pedestrian group ($D$).  This estimation process implies a causal relation of $A \leftrightarrow B$ and $A \leftrightarrow D$ conditioned by $F \rightarrow A$. Certainly, the movement of other road agents also involves complex causal relation reasoning.}
  \label{fig1}
\end{figure}
  \begin{figure*}[!t]
  \centering
 \includegraphics[width=\hsize]{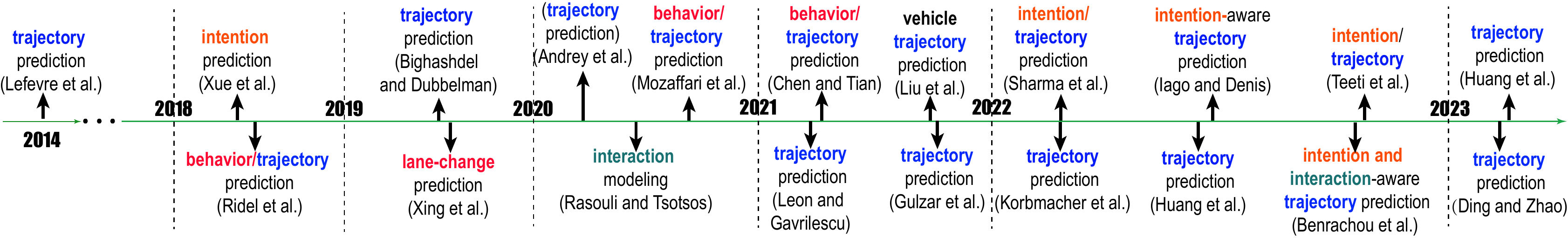}
  \caption{Chronological overview of 20 surveys for the prediction and understanding of intention \cite{xue2018survey,DBLP:conf/ijcai/TeetiKSBC22,DBLP:journals/ijon/SharmaDI22,DBLP:journals/tits/BenrachouGER22,gomes2022review}, behavior \cite{DBLP:journals/tvt/XingLWWACVW19,DBLP:conf/itsc/RidelRLSW18,DBLP:conf/itsc/ChenT21,mozaffari2020deep}, and trajectory \cite{lefevre2014survey,DBLP:conf/itsc/RidelRLSW18,DBLP:conf/itsc/BighashdelD19,rudenko2020human,DBLP:conf/itsc/ChenT21,DBLP:journals/tits/KorbmacherT22,leon2021review,huang2022survey,DBLP:journals/tits/BenrachouGER22,mozaffari2020deep,DBLP:conf/robio/LiuMFZM21,DBLP:journals/access/GulzarMN21,DBLP:journals/ijcv/KongF22,DBLP:conf/ijcai/TeetiKSBC22,DBLP:journals/corr/abs-2302-10463,ding2023incorporating} for pedestrians, vehicles, and their interactions \cite{rasouli2019autonomous,DBLP:journals/tits/BenrachouGER22}.}
  \label{fig2}
\end{figure*}
Based on the denotation, this work terms the ``\textbf{Behavioral Intention Prediction}" (BIP) as the prediction of the intended actions of pedestrians/cyclists or the maneuvers of vehicles (as shown in Fig. \ref{fig1}) under an understanding of surrounding driving scenes. 
However, BIP faces challenges of the accurate understanding of road structure \cite{DBLP:conf/cvpr/MakansiCBB20}, road user interaction \cite{DBLP:journals/tits/FangYQXY22}, moving goal determination, and other prior knowledge understanding, such as the skill, gender, social and cultural factors \cite{rasouli2019autonomous}, \emph{etc.} These clues permeate the social and causal relations, as shown in Fig. \ref{fig1}. The behavioral intention understanding for each agent facilitates the interactive function for autonomous systems \cite{DBLP:journals/scirobotics/X20i}. Designing advanced techniques for predicting the intention of road agents could improve the cognitive level of autonomous systems and help guarantee the safety of all road users. Nowadays, with the vigorous demand for self-driving systems at home and abroad, the corresponding scale of data also grows rapidly, which provides fertile soil for deep learning-based behavioral intention prediction \cite{DBLP:conf/ivs/RasouliYR022,DBLP:journals/tits/GriesbachBH22}. 

\subsection{Distinction to Previous Surveys}
To improve the safety and intelligence of self-driving systems, numerous works have concentrated on the detection, segmentation, and tracking of road agents over the past decades. Some previous surveys \cite{feng2020deep,DBLP:journals/ijcv/LiuOWFCLP20} have comprehensively summarized the pipelines within those fields. 

To clarify the distinction to previous surveys, we extensively searched the related surveys on ``\emph{behavior prediction}", ``\emph{intention prediction}", ``\emph{trajectory prediction}", and ``\emph{crash anticipation}" in Google Scholar.  Fig. \ref{fig2} demonstrates a chronological overview of 20 related high-quality surveys over almost ten years. Trajectory prediction has attracted more attention than other prediction tasks in recent years. More and more research realizes the importance of the intention or interaction of road agents in driving scenes. However, the related surveys are different from our work in the following aspects.

1) Most of the surveys focus on trajectory prediction, and the target of interest is the specific deep learning models \cite{DBLP:conf/itsc/BighashdelD19,DBLP:conf/itsc/ChenT21,huang2022survey}. However, the review of behavioral intention prediction is limited in terms of the key factors (\emph{e.g.}, prediction uncertainty) and the latest progress.

2) The most related surveys to our work are \cite{gomes2022review,DBLP:journals/tits/BenrachouGER22,DBLP:journals/ijon/SharmaDI22}. Among them, these surveys focus on a single type of road agent, such as vehicles \cite{DBLP:journals/tits/BenrachouGER22,gomes2022review}, pedestrians \cite{DBLP:journals/ijon/SharmaDI22}, or human drivers \cite{DBLP:conf/ijcai/TeetiKSBC22}. In addition, the intention types in these surveys are 
monotonous, such as crossing for pedestrians, and lane changing for vehicles. 

Complementary to these surveys, we concentrate on the latest progress in Behavioral Intention Prediction (BIP) for both pedestrians and vehicles, where the trajectory prediction is only an application conditioned by BIP and takes a limited space. What we most want is to extract inspiration for BIP research from the key factors, challenges, and promising models including causality, multimodality, and synthetic-real data collaboration. In addition, we present the most comprehensive review of the available datasets for BIP.

 \begin{figure*}[!t]
  \centering
 \includegraphics[width=0.9\hsize]{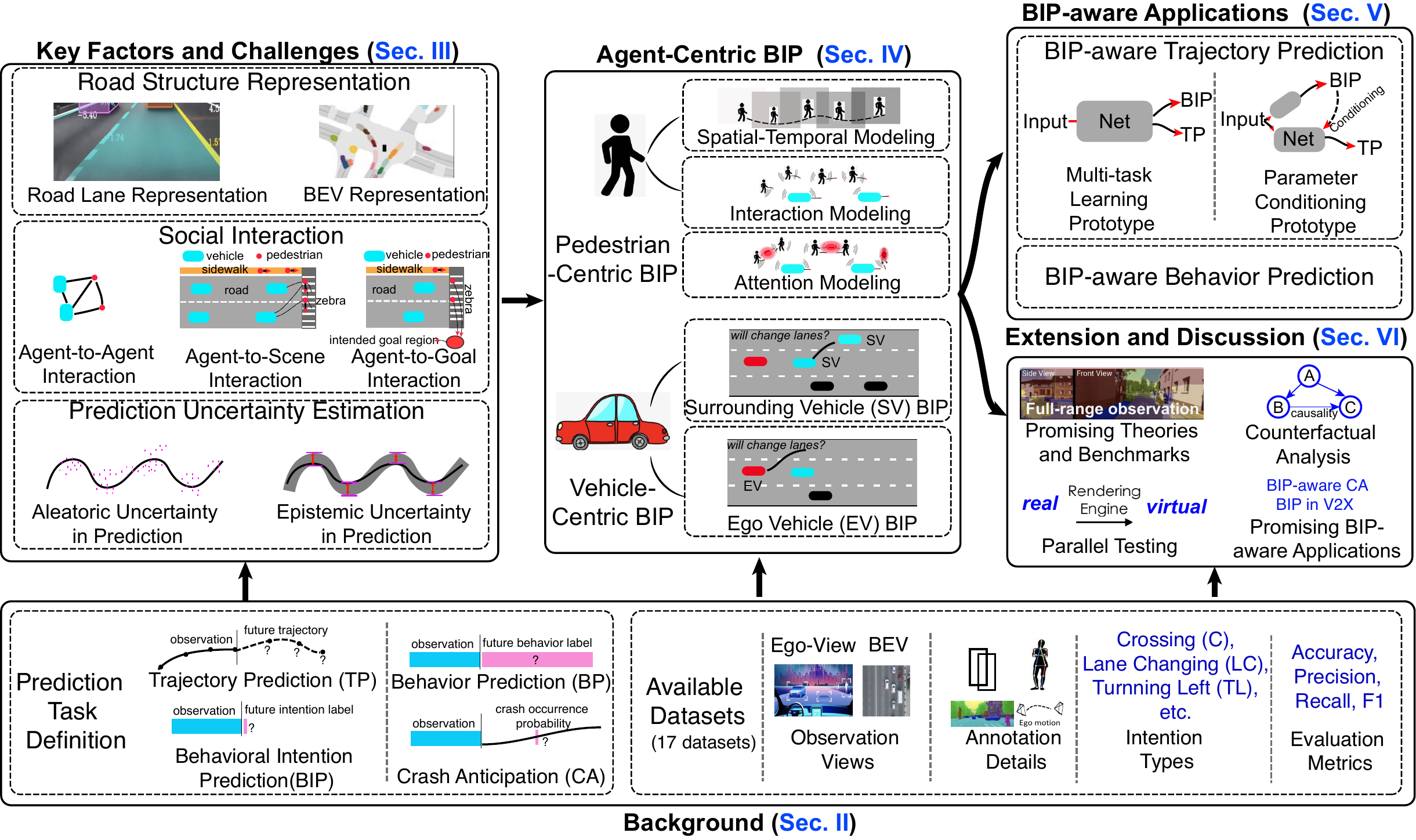}
  \caption{The content taxonomy in this survey. The prediction task definition and available dataset investigation provide a concept basis for this survey. The key factors and challenges are analyzed comprehensively. Then, the agent-centric BIP works for pedestrians and vehicles are chronologically reviewed and discussed. Furthermore, we review the BIP-aware applications from trajectory prediction and behavior prediction, and one-to-one discussions to previous sections are presented for potential issues and possible insights.}
  \label{fig3}
\end{figure*}

\subsection{Taxonomy and Contributions}
This paper reviews the latest works of BIP in driving scenes and presents a full portrait of the \emph{problem definition,  available datasets}, \emph{key factors and challenges}, \emph{agent-centric BIP}, and \emph{BIP-aware applications}. Fig. \ref{fig3} depicts the detailed taxonomy of this survey, where all parts are tightly coupled with clear relations. Based on this work, we want to exhibit the progress of BIP through the research pipeline of \emph{problems and datasets}$\rightarrow$\emph{factors and challenges}$\rightarrow$\emph{approaches}$\rightarrow$ \emph{applications}$\rightarrow$\emph{reseach insights}. The \textbf{contributions} of this survey are as follows.

\begin{itemize}
\item Different from previous surveys, we clarify the definition of different prediction tasks, and provide a more targeted and comprehensive survey on BIP from the available datasets (17 ones), intention types, key factors, approaches, applications, and future insights.
\item The latest progress in behavioral intention prediction is extensively investigated and the new research pipelines are chronologically reviewed and discussed.
\item We provide more discussion for the promising formulations and insights, such as causality, parallel testing, prediction uncertainty modeling, BEV representation, \emph{etc.}
\end{itemize}

The remainder of this work is organized as follows. Sec. \ref{benev} presents the background of the prediction task definition, available datasets, and intention types, which form the concept and denotation basis for the following descriptions. Sec. \ref{ifbi} briefly reviews the key factors and challenges in BIP.  The method progress of agent-centric BIP including pedestrians and vehicles is described and discussed in Sec. \ref{PIM-BIP}. BIP-aware applications including trajectory prediction and behavior prediction are summarized in Sec. \ref{tspre}. Sec. \ref{fcon} presents the discussion for current research and provides potential insights for future research. The conclusion is given in Sec. \ref{fcon1}.

\section{Background}
\label{benev}
Based on the investigation, we find that there is a concept confusion for the tasks of trajectory prediction, behavior prediction, and behavioral intention prediction, and are interchangeably used in this field \cite{DBLP:conf/cvpr/HongSP19,ma2021continual}.  These concepts vary with the output and have different targets of interest. 
\subsection{Prediction Task Definition in Driving Scenes}

 \begin{figure}[!t]
  \centering
 \includegraphics[width=\hsize]{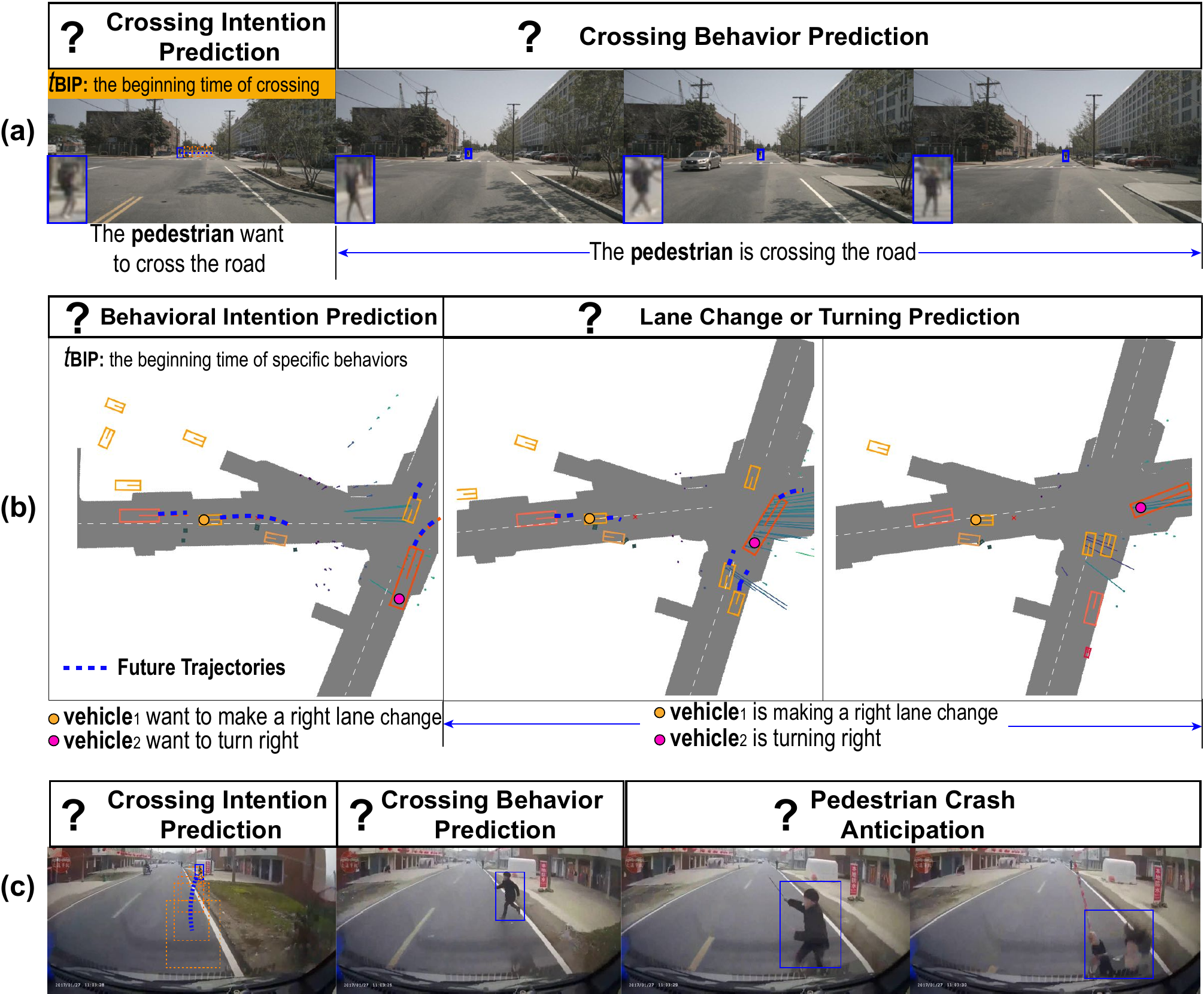}
  \caption{Examples for the behavioral intention prediction, behavior prediction, and crash anticipation, where (a) and (c) captured in the Ego-View are sampled from JAAD \cite{rasouli2017they} and DADA-2000 \cite{DBLP:journals/tits/FangYQXY22} datasets, respectively. (b) denotes the frames of Bird's Eye View (BEV) sampled from the nuScenes dataset  \cite{caesar2020nuscenes}.}
  \label{fig4}
\end{figure}

To begin with, we introduce two observation views in data collection in driving scenes: 1) \textbf{Ego-View}: capturing the video or data point in first-person view (Fig. \ref{fig4} (a)(c)); 2) \textbf{Bird's Eye View (BEV)}: commonly observed in the world coordinate system (Fig. \ref{fig4}(b)). With this setting, we present the terminology for different prediction tasks.

\emph{1) Trajectory Prediction (TP)} is well studied in this field \cite{DBLP:conf/ijcai/TeetiKSBC22,DBLP:journals/corr/abs-2302-10463}, which refers to the process of estimating the future \emph{trajectories} of various entities (vehicles, pedestrians, cyclists, \emph{etc.}) on the road. It analyzes the historical observation states of the entities, \emph{e.g.}, their positions, velocity, and heading, and encodes the information along with surrounding contextual data (\emph{e.g.}, the road map) to predict the future trajectories measured either in the Ego-View or the BEV observation.

\emph{2) Behavior Prediction (BP)} refers to the process of estimating the future \emph{behaviors} of various entities (vehicles, pedestrians, cyclists, \emph{etc.}) on the road \cite{DBLP:journals/neco/PentlandL99}. It can also analyze the historical observation states of the entities, encode them, and predict the \emph{likely} actions (allocated into multiple classes of lane changing, crossing, \emph{etc}.) in the near future, measured by the Ego-View or the BEV).

\emph{3) Behavioral Intention Prediction (BIP)} refers to the process of estimating the \emph{\underline{intended} actions or behaviors} of various entities on the road. It similarly can analyze the historical observation states of the entities, and the contextual information, to infer the \emph{intended} actions of agents in the near future, as shown in the first column of Fig. \ref{fig4} (a-c). 

\emph{4) Crash Anticipation (CA)} refers to the ability to predict and foresee potentially dangerous situations or collisions before they occur on the road. It involves actively analyzing various factors such as the behavior of surrounding vehicles, pedestrians, and road conditions to identify potential hazards or risky situations that may lead to accidents, seeing Fig. \ref{fig4}(c). 

 \begin{table*}[!t]\scriptsize
  \centering
  \caption{Chronological overview of 17 datasets in behavioral intention prediction generated by real data (R) or synthetic data (S) with the \textbf{Intention Types}, \textbf{Annotations}, and the serviceable prediction tasks (\textbf{Pred. Tasks}).}
\begin{tabular}{c|c|c|c|c|c|c}
\toprule[0.8pt]
Datasets &Years/booktitle&Seq. num&Annotations&Intention Types&S/R& Pred. Tasks\\
\hline
\textcolor{magenta}{Daimler} \cite{DBLP:conf/eccv/KooijSFG14} &2014/ECCV &58&I, T, EVV& C, NC& R& BIP, TP\\
\textcolor{magenta}{NGSIM} \cite{NGSIM} &2016/JPO &17,179F*&T, VT, MTV& LLC, RLC, LK& R& BIP, TP\\
\textcolor{magenta}{JAAD} \cite{rasouli2017they} &2017/ICCVW& 346&I, 2DB, W, O, Beh, G, A, BO&C, NC & R&BIP, TP\\
\textcolor{magenta}{HighD} \cite{highDdataset}&2018/ITSC &110 500T*&T, VT, MTV& LLC, RLC, LK& R& BIP, TP\\
\textcolor{magenta}{VIENA2} \cite{aliakbarian2018viena} &2018/ACCV& 15000&I, Beh&S, TL, TR, LLC, RLC, C, NC, W & S&BIP, CA\\
\textcolor{magenta}{INTERACTION} \cite{zhan2019interaction} &2019/arXiv&-&I, 2DB, SS, Beh&MRD, NC, LLC, RLC, M& R& BIP, TP, BP\\
\textcolor{magenta}{PIE} \cite{rasouli2019pie} &2019/ICCV&53&I, 2DB, Beh, EVV&W, STOP, C, NC& R&BIP, BP, TP\\
\textcolor{magenta}{BLVD} \cite{xue2019blvd} &2019/ICRA&654&I, 2DB,T, Beh,3DB&22 types& R& BIP, BP, TP\\
\textcolor{magenta}{PREVENTION} \cite{izquierdo2019prevention} &2019/ITSC&11&I, 2DB,VT, Beh&LLC, RLC, CI, CO& R&BIP, BP\\
\textcolor{magenta}{BPI} \cite{DBLP:conf/itsc/WuWZXW20}&2020/ITSC &120&PO, T & C, NC& R& BIP, BP\\
\textcolor{magenta}{TITAN} \cite{malla2020titan} &2020/CVPR&700&I, 2DB, Beh&ST, C, NC& R&BIP, BP, TP\\
\textcolor{magenta}{STIP} \cite{liu2020spatiotemporal} &2020/IEEE RAL&-&I, 2DB, Beh&C, NC& R& BIP, TP\\
\textcolor{magenta}{PePScenes} \cite{rasouli2020pepscenes} &2020/NeurIPS&850&I, Beh, SS, BO, STD&C, NC& R& BIP, TP, BP\\
\textcolor{magenta}{PSI} \cite{chen2021psi} &2021/arXiv&110&I, 2DB, SS, Beh&C, NC& R& BIP, TP\\
\textcolor{magenta}{LOKI} \cite{DBLP:conf/iccv/GiraseGM0KMC21} &2021/ICCV&664&I, 2DB, 3DB, SS, Age, G, DES, W&C, NC& R& BIP, TP\\
\textcolor{magenta}{Virtual-PedCross-4667} \cite{baijie2022} &2022/ITSC&4667&I, W, 2DB, Beh&C, NC& S& BIP, BP, TP\\
\textcolor{magenta}{DADA-2000} \cite{DBLP:journals/tits/FangYQXY22} &2022/IEEE TITS&2000&I, Beh, DA, W&LLC, RLC, VO, C, NC& R&BIP, DAP, CA\\
\toprule[0.8pt]
  \end{tabular}
    \begin{tablenotes}
\item \scriptsize{ F*: frames. The vehicle trajectory data is collected on the highway with a sampling frequency of 10 Hz. T*: trajectories of vehicles.\\
\textbf{Intention Types:} Crossing (\textbf{C}); Non-Crossing (\textbf{NC}); Walking (\textbf{W}); Standing (\textbf{ST}); Straight Moving (\textbf{SM}); Turning Left (\textbf{TL}); Turning Right (\textbf{TR}); U-Turn (\textbf{UT}); Lane Keeping (\textbf{LK}); Cutting In (\textbf{CI}); Cutting Out (\textbf{CO}); Vehicle Overtaking (\textbf{VO}); Left Lane Changing (\textbf{LLC}); Right Lane Changing (\textbf{RLC}); Stopping (\textbf{STOP}); Pushing (\textbf{P}); Yielding (\textbf{Y}); Merging (\textbf{M}); Moving along the Roundabout (\textbf{MRD}); Accelerating (\textbf{ACCE}), Decelerating (\textbf{DECE}).
\\
\textbf{Annotations:} Image (I); 2D Boxes (\textbf{2DB}); 3D boxes (\textbf{3DB}); Vehicle Type (\textbf{VT}); Ego Vehicle Velocity (\textbf{EVV}); Motion of Target Vehicle (\textbf{MTV}); Driver Attention (\textbf{DA}); Trajectory (\textbf{T}); Weather (\textbf{W}); Behavior (\textbf{Beh}); Pose (\textbf{PO}); Occasions (\textbf{O}); Age (\textbf{Age}); Gender (\textbf{G}); Depth image (\textbf{D}); Human Body Orientation (\textbf{BO}); Destination (\textbf{DES}); Semantic Segments (\textbf{SS}); Scene Text Description (\textbf{STD}).}
\end{tablenotes}
  \label{tab1}
  \end{table*}

From these definitions, we can see that, besides CA and TP, BP and BIP can share a consistent input under the same observation view, while their outputs are different. The output between BP and BIP differs from the ``\emph{\underline{intended}}" word, which means that BIP refers to the conscious and deliberate action or goal \cite{lau2004attention}, and has an earlier timeline than specific behaviors. CA to some extent can integrate the behavioral intention, behavior, and trajectories in certain situations. Different prediction tasks have distinct timelines, and the success of BIP can provide an earlier prompt for safe decisions than other prediction tasks, as shown in Fig. \ref{fig4}(c).

With the clarification of different prediction tasks, we make an overview of the available BIP datasets and intention types. 

\subsection{Available Datasets and Intention Types}
\label{datasets}
For a targeted review, we exhaustively investigate and elaborate on the publicly available datasets for the behavioral intention prediction task. Table. \ref{tab1} presents the attributes of 17 datasets, and the samples are shown in Fig. \ref{fig5}. Almost all pedestrian-centric BIP datasets have pedestrian crossing or not crossing intention. In the following, we describe the main differences among these datasets from the aspects of observation views, annotation details, and intention types.

\subsubsection{Observation Views}
\begin{figure}[!t]
  \centering
 \includegraphics[width=\hsize]{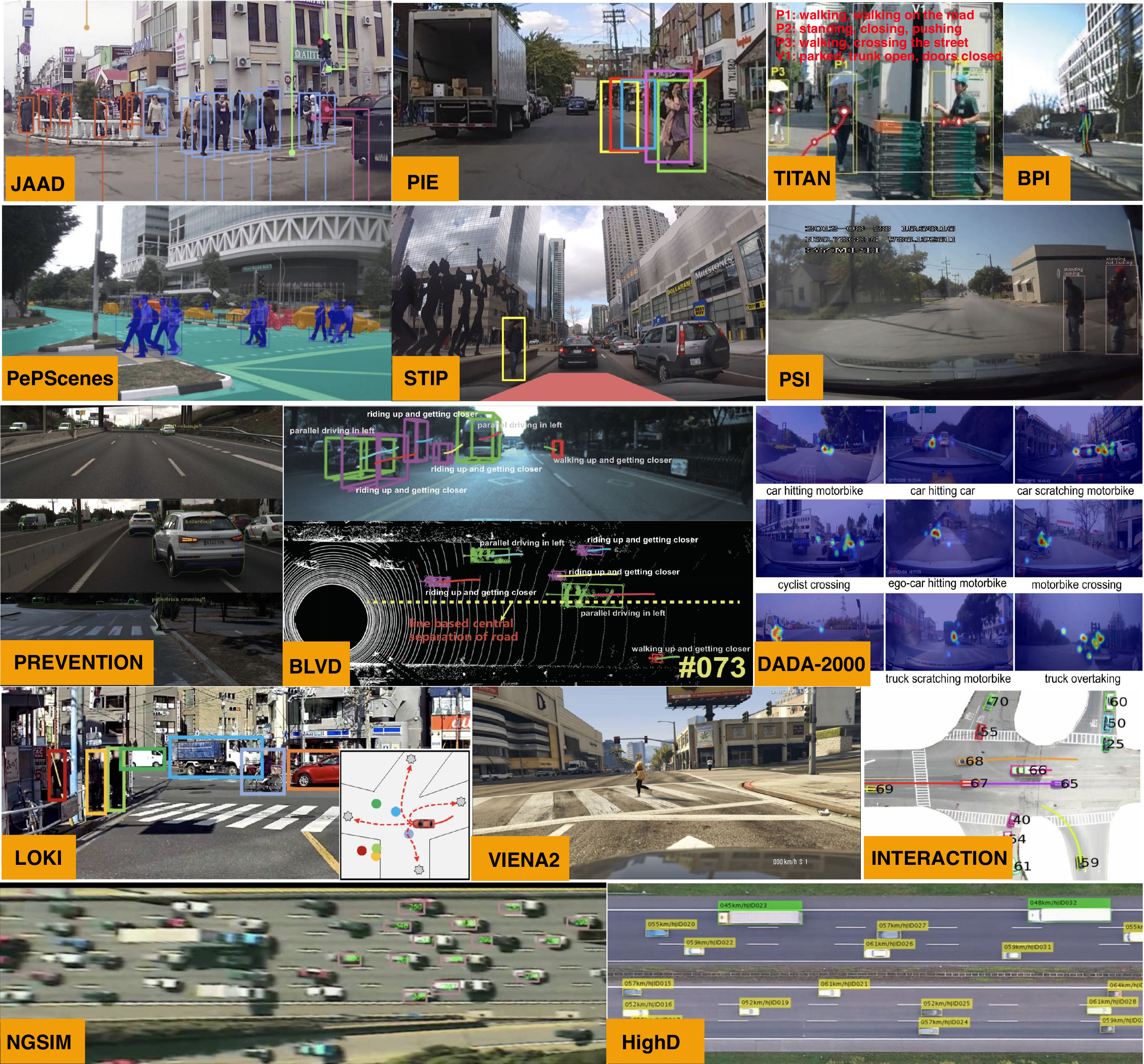}
  \caption{Datasets of JAAD, PIE, TITAN, STIP, BPI, PePScenes, PSI, PREVENTION, BLVD, DADA-2000, VIENA2, LOKI, INTERACTION, NGSIM, and HighD.}
  \label{fig5}
\end{figure}

Observation views have a direct influence on behavioral intention annotation and prediction model designing. \textbf{Daimler} is the pioneering dataset for the prediction of pedestrian crossing or not-crossing, which only has 4 pedestrians in the collection and is captured by gray images. From Fig. \ref{fig5}, the top seven datasets (\emph{i.e.}, \textbf{JAAD} \cite{rasouli2017they}, \textbf{PIE} \cite{rasouli2019pie}, \textbf{TITAN} \cite{malla2020titan}, \textbf{BPI} \cite{DBLP:conf/itsc/WuWZXW20}, \textbf{PePScenes} \cite{rasouli2020pepscenes},  \textbf{STIP} \cite{liu2020spatiotemporal}, and \textbf{PSI} \cite{chen2021psi}) concentrate on the pedestrian-centric BIP with the Ego-View observation, where crossing and not-crossing are two primary intention types. \textbf{TITAN} provides more diverse intention types and supplies the interaction label between pedestrians and road scenes. 
We can see that \textbf{NGSIM} \cite{NGSIM}, \textbf{HighD} \cite{highDdataset}, and \textbf{INTERACTION} \cite{zhan2019interaction} are collected from the BEV view. Compared with the Ego-View, BEV observation can capture a larger spatial range of view and provide a complete movement observation. BEV can provide a good ground-truth verification for the BIP or other prediction tasks. However, the pose and the height of the road agents in Ego-View are clearer. In addition, the Ego-View perception provides the opportunity for collision avoidance by controlling the vehicles in time. Apart from the BEV and Ego-View, the 3D point clouds can also capture the BEV and Ego-View jointly, such as \textbf{BLVD} \cite{xue2019blvd}. However, the raw 3D point cloud has no semantic label or fine-grained pose information of the road agents. Some recent works have been done on capturing the pedestrian point clouds~\cite{zheng2022parameter}, and the panoramic view by multiple cameras, such as the Argoverse 3D dataset \cite{chang2019argoverse} (with seven cameras) or the nuScenes dataset \cite{caesar2020nuscenes} (with six cameras). However, these 3D point cloud datasets do not provide the behavioral intention label. Therefore, in the future, these panoramic view datasets can be extended with behavioral intention or behavior labeling.

\subsubsection{Annotation Details}
The annotation details in these datasets are intricate and provide different modeling and insights for behavioral intention prediction. The object bounding boxes, trajectories, road entities, driver attention, long-term intended goal regions, risk levels, age, and gender of pedestrians are all important for the safety evaluation in the driving scene. \textbf{PSI} \cite{chen2021psi} provides the Scene Text Description (STD) for different situations, which presents another perspective for scene understanding. From Fig. \ref{fig5}, we can see that only \textbf{DADA-2000} considers the crash scenarios. In Table. \ref{tab1}, we present 19 kinds of annotation attributes. If we want to check the counterfactual analysis for the BIP on crash anticipation or collision avoidance, DADA-2000 will be feasible. In addition, with the development of text-to-video diffusion models \cite{an2023latent,DBLP:journals/corr/abs-2212-11565}, some attempts to create editable driving scenes may be promising. In addition, crow-view annotation (\emph{e.g.}, the calibration between Ego-View and BEV) is another direction for the cross-validation or counterfactual analysis of BIP.

\subsubsection{Intention Types}
Different datasets focus on distinct intention types. In Table. \ref{tab1}, we mainly summarize 21 common intention types. Most of the datasets only contain the Crossing (C) or Not Crossing (NC) of pedestrians. Based on the road structure, TITAN \cite{malla2020titan} provides fine-grained pedestrian behavior labels, such as \emph{Pushing} (P), or \emph{Standing} (ST). The label in TITAN is actually a sequential action label for each road agent. 
As for vehicle-centric intention types, most of them (NGSIM \cite{NGSIM} and HighD \cite{highDdataset}) provide the Lane Changing (LC) intention, and \textbf{PREVENTION} \cite{izquierdo2019prevention} has the \emph{``Cutting In"} (CI) and \emph{``Cutting Out"} (CO) intention, in which ``risk levels" are provided for the vehicle Lane Changing (LC) intention. Because there are round paths, INTERACTION \cite{zhan2019interaction} offers the behavioral intention of \emph{``Moving along the Roundabout"} (MRD).
In addition, BLVD annotates 22 fine-grained types of intention, including 12 types, 7 types, and 8 types of behavioral intention for vehicles, pedestrians, and riders (cyclists or motorbikes), respectively. It is promising for fine-grained BIP in the driving scene. Concisely, the fine-grained intention types in BLVD can be viewed in \cite{xue2019blvd} for details.
Based on the comparison, we can see that the behavioral intention types for different road agents in most current datasets are far from being meticulous. Some safe-critical behavioral intention types, such as \emph{``vehicle running conversely"} and \emph{``braking"}, \emph{etc.}, and many kinds of intention types involving the interactions between different road agents with road entities (\emph{e.g.}, the sidewalk, bus station, and steep slope, \emph{etc.}) are not exploited. 

\subsubsection{Evaluation Metrics}
Most of the BIP works in this field determine the future intention as a classification problem. Consequently, the \textbf{Accuracy (Acc)}, \textbf{Precision (Pre)}, \textbf{Recall (Rec)}, and \textbf{F1}-measure (\textbf{F1}) are four common metrics for evaluation, where the computing methods are:
Acc=$\frac{TP+TN}{TP+TN+FP+FN}$, Pre=$\frac{TP}{TP+FP}$, Rec=$\frac{TP}{TP+FN}$, and F1=$2\times \frac{Pre\times Rec}{Pre+Rec}$. $TP$ and $FP$ are respectively the predicted positive intention samples in the true positive and negative set, and $TN$ and $FN$ are respectively the predicted negative intention samples in the true negative and positive set.

With the definition of the prediction tasks and the introduction of datasets and intention types, we describe the key factors, challenges, promising models, and BIP-aware applications in the following, where the dataset name, intention types, and annotations are consistent throughout the whole paper.

\section{Key Factors and Challenges}
\label{ifbi}
The mixed traffic scene makes the factors complex for Behavioral Intention Prediction (BIP) \cite{rasouli2019autonomous, guo2019safe}. BIP of surrounding agents needs to consider the agent type, road structure, social relation, action tendency, and intention types, which consist of factors from robust road structure representation, social interaction modeling, and prediction uncertainty estimation, resulting in various challenges for BIP research.

\subsection{Road Structure Representation}
The road scene is a highly structured environment, and the road structure contains consistent traffic rules. In the meantime, road users should obey the road etiquette \cite{DBLP:conf/cvpr/HongSP19,belkada2021pedestrians}. All the static road entities (road lanes, road boundary, \emph{etc.}) and dynamic agents (pedestrians, vehicles, \emph{etc.}) in the road scene constitute the contextual information for safe driving \cite{DBLP:journals/ftrob/GargSDMCCWCRGCM20}. Therefore, the first kind of key factor is the road structure representation, which is often jointly modeled by the static road entities and dynamic agents.

The most universal road agents in the mixed road scene are pedestrians, cyclists, motorbikes, buses, trucks, cars, trailers, \emph{etc.} Based on the movement patterns, the road layout, and observation views (\emph{i.e.}, Ego-View, and BEV), the intention of interest is different. The agent-centric intentions, such as \emph{``crossing"}, \emph{``walking"}, \emph{``running"}, \emph{``stopping"} for pedestrians, or \emph{``lane keeping"}, \emph{``turning left or right"}, \emph{``braking"}, \emph{``accelerating"}, \emph{``lane changing"} for vehicles, usually correlates with the specific situations. 
\subsubsection{Road Lane Representation}
The road lane is a primary clue for the road structure representation and is of great interest to the autonomous driving community \cite{sato2022towards,feng2022rethinking,paek2022k,li2022multi}. In this field, most researches detect the road lanes in camera videos \cite{DBLP:journals/tits/AndradeBFSNMOFT19} or differentiate the reflectance of the scanned targets by 3D-LiDAR \cite{DBLP:journals/tvt/TaoHZXZ22}. Using lane centerlines as the anchors for constraining the trajectory prediction is widely investigated \cite{DBLP:conf/cvpr/NarayananMPLC21,DBLP:conf/eccv/LiangYHCLFU20}. Lane graph representations \cite{DBLP:conf/iros/ZengLLU21,DBLP:conf/eccv/LiangYHCLFU20} are modeled from raw map data to explicitly explore the complex road topology and long-range dependencies, where there are three types of interaction between agents and the road map (\emph{i.e.}, lane-to-lane, lane-to-agent, agent-to-lane) in the lane graph representation. For an in-depth utilization of the information on road lanes, Hong \emph{et al.} \cite{DBLP:conf/cvpr/HongSP19} unify the representation which encodes the road map in a spatial road grid, allowing the use of fusing complex scene context of entity-entity and entity-environment interactions. However, these lane graph representations only consider the road boundaries or the lane centerlines of the High-Definition (HD) maps built in advance. As for behavioral intention prediction, road lane-based representation only can be reflected on the Lane Changing (LC) intention or Lane Keeping (LK) intentions.
 \begin{figure}[!t]
  \centering
 \includegraphics[width=\hsize]{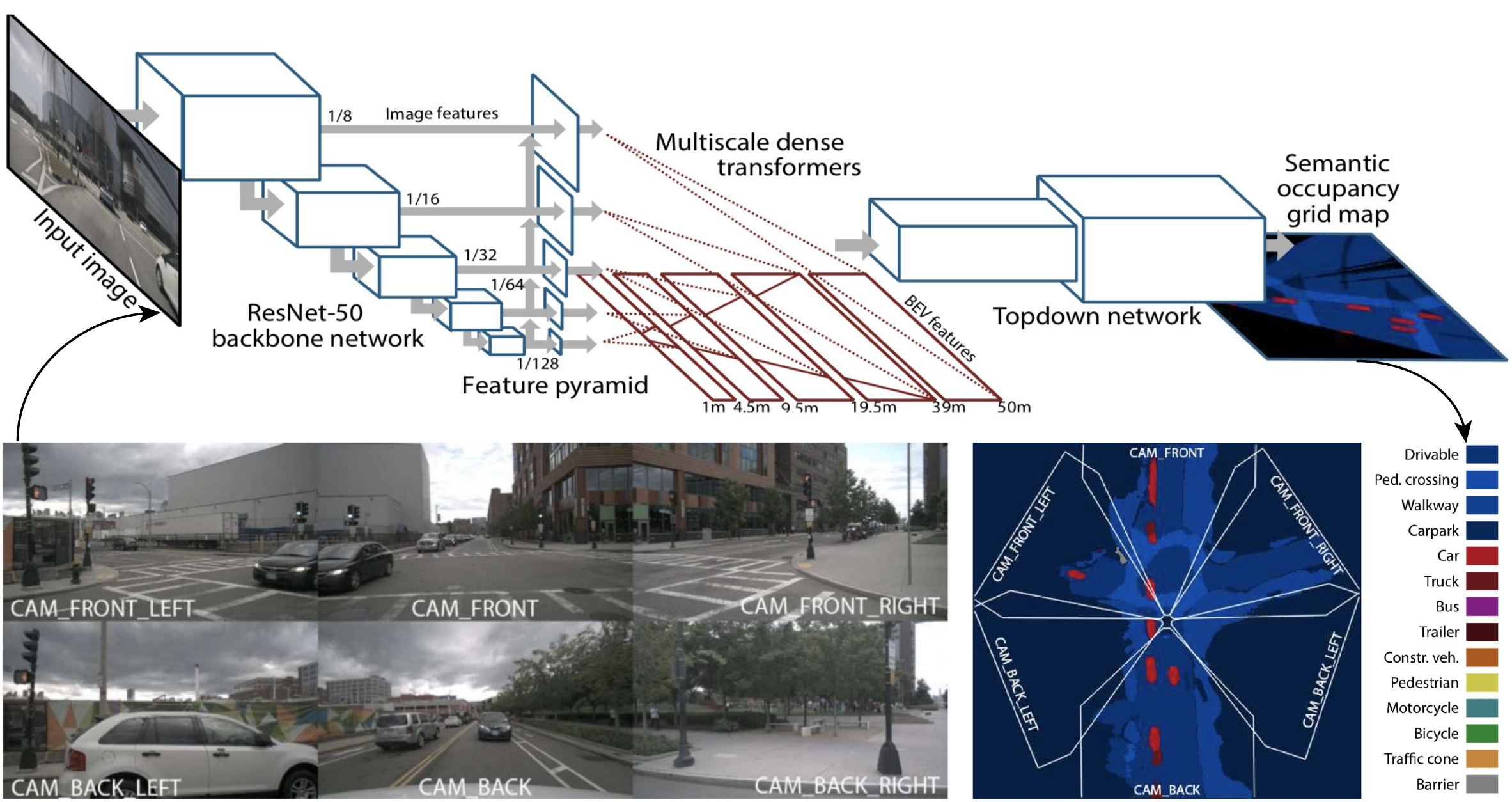}
  \caption{The BEV representation with deep feature learning \cite{DBLP:conf/cvpr/RoddickC20}. }
  \label{fig6}
\end{figure}

\subsubsection{BEV Representation}
Because HD maps commonly need to be built in advance with high cost, some attempts concentrate on the road topology representation with on-board camera data \cite{DBLP:journals/corr/abs-2112-10155,DBLP:conf/iccv/HomayounfarLMFW19}. This gives rise to the road structure representation by Bird’s Eye View (BEV) \cite{zhou2022cross,saha2022pedestrian,zhao2022scene} (as shown by Fig. \ref{fig6}) transferred from the raw camera videos by deep feature transformation \cite{DBLP:conf/cvpr/RoddickC20}. The BEV view shows a clearer road layout than that captured by the Ego-View, where the scale of the object, physical attributes of movement (\emph{e.g.}, object velocity and orientation, \emph{etc.}), and the social interaction can be perceived without geometric distortion. However, BEV representation relies on the inter-correlation between the camera videos and the road map grid. The inter-correlation is easily influenced by the perception distortion in camera videos. Sometimes, the lane extraction in camera videos is challenging \cite{DBLP:journals/tits/CuiXZ16,li2021bsp}. Consequently, the current paradigm commonly denotes BEV representation by the intermediate layers of deep features \cite{DBLP:conf/eccv/LiWLXSLQD22,hu2023planning} which provides richer road structure representation than road lane representation.

\subsection{Social Interaction Modeling}

Social interaction modeling is the key factor for driving scene perception all the time. Previous works \cite{rasouli2019autonomous,DBLP:conf/cvpr/AlahiGRRLS16,DBLP:conf/icra/0002GLU20,DBLP:conf/eccv/SadatCRWDU20,DBLP:journals/corr/abs-2101-02385,DBLP:conf/cvpr/0002SU21,DBLP:conf/corl/CasasLU18} involve the interaction or relation of the road agents to fulfill the future state prediction of agents. The intermediate feature representation of the occupancy map taken by agents \cite{DBLP:conf/eccv/SadatCRWDU20,DBLP:conf/cvpr/0002SU21} is a common choice for interaction modeling in intention or trajectory prediction. Actually, the interaction \cite{DBLP:journals/tits/KorbmacherT22} provides the prior for potential moving tendency, such as the reachability prior \cite{DBLP:conf/cvpr/MakansiCBB20}.

\subsubsection{Agent-to-Agent Interaction}

There are a vast number of models for agent-to-agent interaction in this field, and the social-LSTM \cite{DBLP:conf/cvpr/AlahiGRRLS16} is the pioneering model for future state prediction of agents. Inspired by this, many variants of the social-LSTM-based interaction \cite{DBLP:conf/cvpr/ZhangO0XZ19,wang2022social,rudenko2020human} have boomed quickly since 2016. Each agent in Social-LSTM is modeled as an individual LSTM and shares the social relation by a social pooling system, which has a promising ability for modeling the temporal dependence of agent state but is limited when meeting crowd agents. For this issue, the Graph Neural Networks (GCN) can fulfill a flexible social relation modeling \cite{DBLP:conf/cvpr/MohamedQEC20,su2022trajectory,DBLP:journals/corr/abs-2010-05507,DBLP:conf/itsc/LiYC19,DBLP:conf/nips/KosarajuSM0RS19,choi2020drogon}. Each node in the graph can represent diverse information of locations, velocity, agent types, \emph{etc.} However, GCN-based methods need the number of agents to be temporally consistent, which is limited in highly dynamic driving scenes. 
\subsubsection{Agent-to-Scene Interaction}
For a long time, the road map or the occupancy map is encoded with a dense \emph{rasterized} processing, which has been adopted in many popular trajectory prediction methods, such as DESIRE \cite{DBLP:conf/cvpr/LeeCVCTC17}, IntentNet \cite{DBLP:conf/corl/CasasLU18}, CoverNet \cite{phan2020covernet}, Trajectron++ \cite{DBLP:conf/eccv/SalzmannICP20}, MultiPath \cite{DBLP:conf/corl/ChaiSBA19}, Target Driven Trajectory (TNT) \cite{DBLP:conf/corl/ZhaoGL0SVSSCSLA20}, and so on. These methods typically encode the road map with Convolution Neural Networks (CNN), while the structure of the road layout is not modeled well with the restricted perception field of CNN. MultiPath++ \cite{DBLP:conf/icra/VaradarajanHSRN22} extends MultiPath with an efficient polyline encoding for agent-to-scene relations, which exploits the region-to-region relation for a better prediction ability. However, the polyline representation seems difficult requiring accurate annotation. 

\begin{figure}[!t]
  \centering
 \includegraphics[width=\hsize]{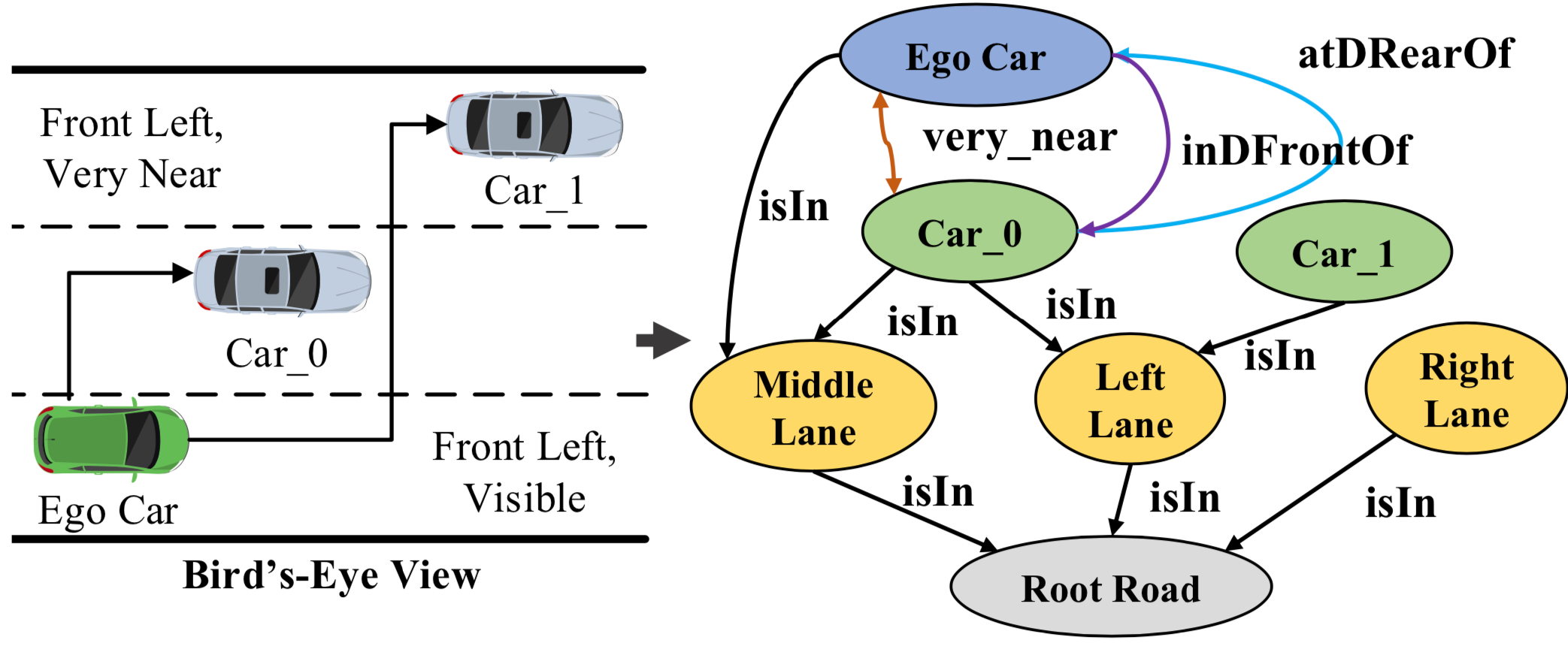}
  \caption{The scene graph generation by RoadScene2Vec \cite{malawade2022roadscene2vec}. }
  \label{fig7}
\end{figure}

Agent-to-scene interaction shows a promising constraint for reducing the \emph{implausible} trajectories \cite{DBLP:conf/eccv/ParkLSBKFJLM20}. The aforementioned agent-to-scene interactions need to pre-annotate or build the road map effectively, which limits flexibility in various situations when the road map information is not accurately obtained. Recently, for agent-to-scene interaction, the scene graph has attracted more attention \cite{fang2023heterogeneous,DBLP:journals/corr/abs-2304-05277}. The scene graph can reflect the common social interaction knowledge via large-scale data learning from raw video or point cloud data, and can be vectorized effectively, such as the RoadScene2Vec \cite{malawade2022roadscene2vec}, as shown in Fig. \ref{fig7}. Under this insight, Song \emph{et al.} \cite{DBLP:conf/iros/SongKZWM022} vectorize the traffic scene graph in each frame in pedestrian crossing prediction with improved performance.

\subsubsection{Agent-to-Goal Interaction}
\label{goal} 
\begin{figure}[!t]
  \centering
 \includegraphics[width=\hsize]{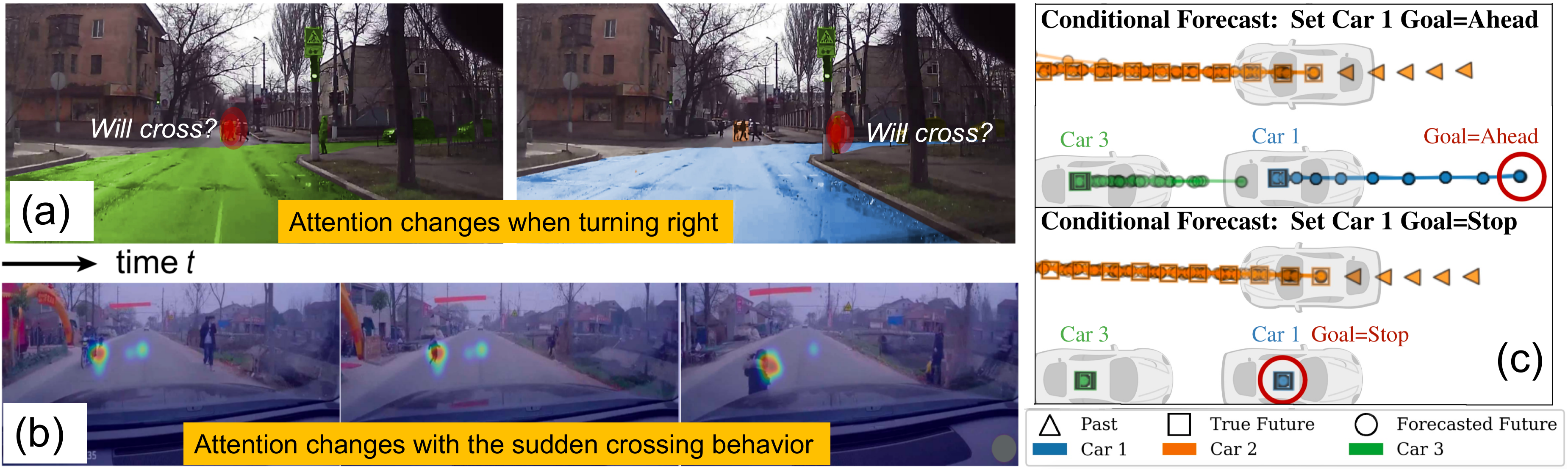}
  \caption{The goal-centric scene representation and prediction, where (a) denotes an attention change (credits to \cite{rasouli2018joint}),  (b) is a sample for driver attention evolution in accident scenario (credits to \cite{DBLP:journals/tits/FangYQXY22}), and (c) specifies a goal-centric vehicle intention prediction (credits to \cite{rhinehart2019precog}). }
  \label{fig8}
\end{figure}

Along with the driving scenes, BIP is commonly influenced by different intended goal areas. With this in mind, we can see that the importance of the road agents is different and changes with the varying driving scenes \cite{ohn2017all,rasouli2018joint}, as shown in Fig \ref{fig8}(a). In this community, the \emph{driver attention} is a direct clue to reflect the important and preferred goals. As shown in Fig. \ref{fig8}(b), the intended fixation of drivers not only reflects where the drivers want to go but can also help to discover dangerous objects \cite{DBLP:journals/tits/FangYQXY22}. 

For the agent-to-goal interaction, the ``\emph{goal}" is commonly modeled as the future intended region defined as destination coordinates \cite{DBLP:conf/cvpr/LeeCVCTC17,DBLP:conf/cvpr/ChiaraCDCCB22}. If the goals are pre-known, the goal-conditioned prediction can be inferred by the inverse optimal control \cite{DBLP:conf/eccv/KitaniZBH12} or inverse reinforcement learning \cite{DBLP:conf/aaai/ZiebartMBD08}. However, for future agent state prediction, the coordinate-based goals may be unknown beforehand. Therefore, the agent-to-goal interaction is dynamic with the behavior and location changing of the agents, and the intended goal estimation is important. Here, the agent-to-scene interaction once again plays a key role in restraining the possibly intended goals of the agents \cite{DBLP:conf/cvpr/ChiaraCDCCB22}. For example, map-adaptive goal path \cite{DBLP:conf/corl/ZhangSHHM20} generates a set of possible goal-directed future path anchors by the road lane constraint. The Goal Area Network (GANet) \cite{wang2022ganet} models the possible goal areas rather than the exact goal coordinates for motion prediction. Within GANet, the possible goals are estimated by calculating the loss between the inferred goal locations with the endpoint of the ground-truth trajectories. Commonly, these agent goal estimation works need to pre-define many goal anchors (candidate goal coordinates) and conduct heuristic or rule-based goal selection. Apparently, the quality of the goal anchors has a heavy impact on the prediction accuracy and a wrong estimation may directly change the future intention in prediction. Target-driveN Trajectory (TNT)  \cite{DBLP:conf/corl/ZhaoGL0SVSSCSLA20} and DenseTNT \cite{DBLP:conf/iccv/Gu0Z21} are two popular models with the goal estimation of ego-vehicles, where DenseTNT estimates the probabilities of goal coordinate candidates relaxing the requirement of the heuristic future path anchors.  

The aforementioned interaction categories correlate with each other. As for the BIP, the highly  socialized driving scenes permeate various interactions but are challenging because of the dynamic behaviors of agents, frequent disappearance or the emergence of new objects, and complex road structures.

\subsection{Prediction Uncertainty Estimation}

``\emph{It is far better to foresee even without certainty than not to foresee at all.}"
--Henri Poincare,  \emph{Foundations of Science} \cite{poincare-royce-2014}. 

The inherent multi-modality, partial observability, short time scales, data limitation, intention type imbalance \cite{westny2021vehicle}, domain gap~\cite{zheng2021rectifying}, and deficiency can all cause uncertainty. In addition, because of the generalizability of deep learning models, the predicted behavioral intention distribution may involve bias. There are two kinds of uncertainties: 1) the \emph{aleatoric uncertainty} also termed as observation uncertainty, that refers to the inherent randomness or variability that is presented in a system or process, and 2) the \emph{epistemic uncertainty}, also called model uncertainty, that is arisen by the limited knowledge or information about a system or process. 

In particular, aleatoric uncertainty refers to the irreducible, objective, or stochastic uncertainty of a physical system (sensor ability) or environment (severe weather, low light condition, \emph{etc.}), and cannot be reduced even with complete knowledge or understanding of the underlying factors \cite{DBLP:conf/nips/KendallG17}. On the contrary, epistemic uncertainty accounts for uncertainties in the model parameters and can be reduced through improved data collection, improved modeling techniques, or increased knowledge about the system.  
\subsubsection{Aleatoric Uncertainty in Prediction}
In agent state prediction, the ways for these two types of uncertainties are different. For example, in order to weaken the aleatoric uncertainty, extra clues, such as High-Definition Map (HD Map), Birds' Eye View (BEV), \emph{etc.}, are taken into account for future prediction. Full-range BEV representation is adopted in a recent work StretchBEV \cite{akan2022stretchbev} for the future instance prediction, and stretches the spatial scene for longer time horizons than previous works. MultiPath \cite{DBLP:conf/corl/ChaiSBA19} proposes multiple probabilistic anchor trajectory hypotheses with the aid of HD Map, and models the future state as a Gaussian Mixture Model (GMM), where an \emph{intention uncertainty} is defined for inferring the latent coarse-scale intention or desired goals. Yalamanchi \emph{et al.} \cite{DBLP:conf/itsc/YalamanchiHHD20} address the long-term future prediction with the uncertainty-aware trajectories with lane-based paths. In order to model the aleatoric uncertainty, various kinds of probability models are developed, such as the Gaussian model \cite{DBLP:conf/icra/IvanovicLSCP22,DBLP:conf/eccv/SalzmannICP20}, GMM \cite{DBLP:conf/corl/Jain0LXFSU19,DBLP:conf/iccv/Zheng0ZTN0021}. Actually, because of the dynamic and objective intention, Gaussian distribution usually expresses the scene sensitivity poorly and the inherent multimodal nature of the future road agent states increases the uncertainty. For example, the pedestrian may continue along a sidewalk or cross a crosswalk, as shown in Fig \ref{fig9} (a).

\subsubsection{Epistemic Uncertainty in Prediction}
For epistemic uncertainty, various models introduce multiple kinds of information or prior knowledge to reduce the prediction uncertainty. For example, based on the interaction nature aforementioned, different road agents can also raise \emph{collaborative uncertainty} (CU) \cite{DBLP:journals/corr/abs-2207-05195} because of the dynamics of the interaction. The consideration of CU enables to evaluation of the interaction uncertainty in the multi-modal state prediction. In addition to the model uncertainty, cross-dataset domain adaptation is also a helpful way for introducing the data knowledge in source datasets to the target dataset. For example, Gesnouin \emph{et al.} \cite{DBLP:conf/ivs/GesnouinPSM22} investigate the cross-dataset generalization for pedestrian crossing intention prediction, and find that the dataset shift degrades the quality of predictions regardless of the model selection, even with well-calibration on the training and testing distributions in each dataset. Deep ensembles of multiple networks seem to be beneficial for boosting the model performance under data shift \cite{ovadia2019can}.
 \begin{figure}[!t]
  \centering
 \includegraphics[width=\hsize]{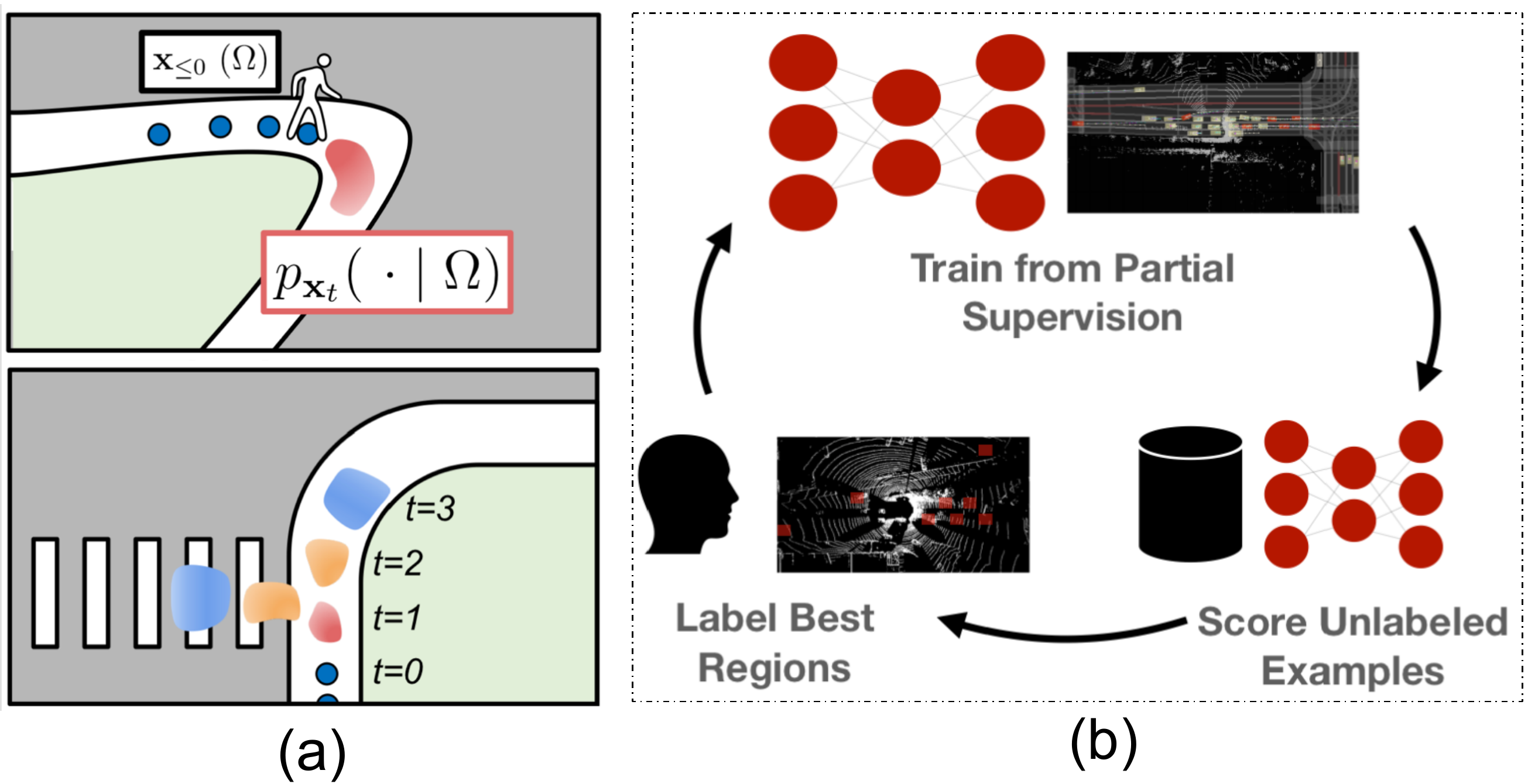}
  \caption{The aleatoric uncertainty with (a) Gaussian distribution for future paths \cite{DBLP:conf/corl/Jain0LXFSU19} and (b) the epistemic uncertainty with partially labeled data \cite{segal2022just}. }
  \label{fig9}
\end{figure}

\textbf{What uncertainties do we need to consider for BIP?} Although there is little work for this question, we can seek the answer from the work on \emph{Bayesian} deep learning \cite{gawlikowski2021survey,itkinainterpretable} in computer vision \cite{DBLP:conf/nips/KendallG17}. Aleatoric uncertainty can be focused on when we have sufficient data or with real-time demand, and epistemic uncertainty is important when we encounter safety-critical applications with small data.  As for BIP in the driving scene, the intention types of road agents are multitudinous. Consequently, it is impossible to collect enough data in practical use for each type of behavioral intention and may involve many types with small scale of samples.  In the meantime, each agent in the driving scene may have different intentions at each time step, which implies natural aleatoric uncertainty. 
Furthermore, with the influence of partial observation, \emph{few-shot} \cite{zang2021few} or \emph{zero-shot} \cite{DBLP:conf/emnlp/XiaZYCY18} learning models with limited labels or important labels \cite{segal2022just} (as shown in Fig. \ref{fig9}(b)) can also be taken to address epistemic uncertainty. Human-machine hybrid intelligence will have an important role in future prediction with the humans' help to correct prediction errors \cite{styles2019forecasting} in an active learning setting.

\subsection{In Summary}
The factors in BIP mainly have an essential impact on model design. We make a summary for this section. 

1) The observation range of the Ego-View is limited, where the prediction of crossing and not-crossing intentions is the primary task. 
Therefore, the observation views will become unified with the development of BEV representation for autonomous driving systems because of the full range and clear observation of surrounding scenes. Currently, BEV observation is transformed from raw camera videos with deep feature learning. However, the performance of fine-grained road entities (static entities and dynamic agents) projection in BEV representation still needs to be enhanced. 

2) The interaction types aforementioned will coexist all the time, which relies on specific driving scenarios, such as highways, urban roads, \emph{etc}. Agent-to-agent interaction relies on accurate agent attributes. To be concise and effective, agent-to-agent interaction needs to consider the neighborhood range and the importance of the agents because there is only a very small proportion of the target agents that have influences on the ego agent. Agent-to-scene and agent-to-goal interaction need accurate road structure representations, which need to pre-construct the HD map in LIDAR sensors and are easily influenced by the weather and light conditions if the BEV representation transformed by cameras is used. 

3) The main issue for prediction uncertainty is the data shift and intention type imbalance. Recently, Digital Twining (DT) \cite{xie2021digital} or Parallel Intelligence (PI) \cite{wang2016steps} may be promising for epistemic uncertainty by generating large-scale behavioral intention data in long-tailed and critical situations.

\begin{table*}[!t]\footnotesize
  \centering
  \caption{Chronological overview of the pedestrian-centric BIP methods, \emph{w.r.t.}, Years, Inference Models, Multimodality Fusion Strategies, Clue Types, and Intention Types.}
  \renewcommand{\arraystretch}{1}
     \setlength{\tabcolsep}{0mm}{
\begin{tabular}{c|c|c|c|c|c|c|c}
\toprule[0.8pt]
Ref. &Years&booktitle&Inference Models&Clue Types&Fusion Strategies&Intention Types& Datasets\\
\hline
Volz \emph{et al.} \cite{DBLP:conf/itsc/VolzBMGSN16} &2016&ITSC& CNN, LSTM, SVM& I& concat & C, NC& Self-c\\
\hline
Saleh \emph{et al.} \cite{DBLP:journals/tiv/SalehHN18} &2018&IEEE TIV&LSTM& T& concat &C, NC&Daimler\\
\hline
Fang \emph{et al.} \cite{DBLP:conf/ivs/FangL18} &2018&IV& CNN&I, PO, 2DB& concat & C, NC&JAAD\\
\hline
Ghori \emph{et al.} \cite{DBLP:conf/ivs/GhoriMBBDDO18} &2018&IV& CNN, LSTM&I, PO, 2DB& concat &C, NC&Daimler\\
\hline
Varytimidis \emph{et al.} \cite{DBLP:conf/sitis/VarytimidisADE18} &2018&SITIS&\makecell{SVM, ANN, kNN,\\Decision Trees}&I, 2DB, EVV, BO& concat &C, NC&JAAD\\
\hline
Rasouli \emph{et al.} \cite{DBLP:conf/bmvc/RasouliKT19} &2019&BMVC&GRU&I, PO, 2DB, EVV& concat &C, NC&PIE\\
\hline
Rasouli \emph{et al.} \cite{DBLP:conf/iccv/RasouliKKT19} &2019&ICCV&LSTM&I, 2DB& concat &C, NC, W&PIE, JAAD\\
\hline
Gujjar \emph{et al.} \cite{DBLP:conf/icra/GujjarV19} &2019&ICRA&\makecell{Residual Encoder-Decoder,\\3DCNN}&I&concat&C, NC&JAAD\\
\hline
Saleh \emph{et al.} \cite{DBLP:conf/icra/SalehHN19} &2019&ICRA& \makecell{Spatial-Temporal (ST) \\DenseNet}&I& w/o fusion&C, NC&JAAD\\
\hline
Cadena \emph{et al.} \cite{DBLP:conf/itsc/CadenaYQW19} &2019&ITSC& GCN&PO& concat &C, NC&JAAD\\
\hline
Piccoli \emph{et al.} \cite{DBLP:conf/acssc/PiccoliBPSNTABR20} &2020&ACSSC&STDenseNet&I, PO, 2DB&concat &C, NC&JAAD\\
\hline
Bouhsain \emph{et al.}\cite{DBLP:journals/corr/abs-2010-10270} &2020&hEART&LSTM&2DB, EVV& concat &C, NC&JAAD\\
\hline
Fang \emph{et al.} \cite{DBLP:journals/tits/FangL20} &2020&IEEE TITS &CNN&I, PO, 2DB& concat & C, NC, TL, TR, STOP&JAAD\\
\hline
Liu \emph{et al.} \cite{liu2020spatiotemporal} &2020&IEEE RAL&GCN&I, 2DB& concat & C, NC&JAAD, STIP\\
\hline
Wu \emph{et al.} \cite{DBLP:conf/itsc/WuWZXW20} &2020&ITSC&\makecell{LSTM, Dynamic \\Bayesian Network}&I, PO& concat & C, NC&BPI\\
\hline
Alvarez \emph{et al.} \cite{DBLP:conf/ivs/Morales-Alvarez20} &2020&IV&GRU &I, 2DB, PO, EVV& concat &C, NC&PIE\\
\hline
Kotseruba \emph{et al.} \cite{DBLP:conf/ivs/KotserubaRT20} &2020&IV& Logistic Regression Classifier&I, 2DB, EVV& concat &C, NC&PIE\\
\hline
Chaabane \emph{et al.} \cite{chaabane2020looking} &2020&WACV&C3D, Conv-LSTM&I& w/o fusion &C, NC&JAAD\\
\hline
Cao and Fu \cite{20204009255721} &2020&Journal of Physics& GCN, Conv-LSTM&I, 2DB, PO, EVV& concat &C, NC, ST&PIE\\
\hline
Rasouli \emph{et al.} \cite{DBLP:conf/iccv/RasouliRL21} &2021&ICCV&LSTM&T, I, 2DB, EVV, SS& concat &C, NC&JAAD, PIE\\
\hline
Chen \emph{et al.} \cite{DBLP:conf/iccvw/ChenTD21} &2021&ICCVW&LSTM, GCN&I, 2DB, PO& concat & C, NC&PIE\\
\hline
Chen \emph{et al.} \cite{chen2021psi} &2021&arxiv&LSTM, Conv-LSTM&I, 2DB, PO& attentive fusion & C, NC&PSI\\
\hline
Lorenzo \emph{et al.} \cite{lorenzo2021intformer} &2021&arxiv&Transformer&I, 2DB, PO, EVV& concat & C, NC&PIE, JAAD\\
\hline
Singh \emph{et al.} \cite{DBLP:conf/iccvw/SinghS21} &2021&ICCVW&3DCNN&PO, 2DB& concat &C, NC&JAAD\\
\hline
Yau \emph{et al.} \cite{DBLP:conf/icra/YauMRLRL21} &2021&ICRA&STGCN, LSTM&I, 2DB, EVV& concat &C, NC&PePScenes\\
\hline
Neogi \emph{et al.} \cite{DBLP:journals/tits/NeogiHDYD21} &2021&IEEE TITS&Factor-CRF&I, SS, D, EVV& concat & C, NC&JAAD\\
\hline
Yao \emph{et al.} \cite{DBLP:conf/ijcai/YaoAJV021} &2021&IJCAI&CNN, MLP&I, 2DB, SS& concat & C, NC&PIE, JAAD\\
\hline
Razali \emph{et al.} \cite{20212910650116} &2021&TRC &3DResNet50&I, PO&attentive fusion& C, NC&JAAD\\
\hline
Kotseruba \emph{et al.} \cite{kotseruba2021benchmark} &2021&WACV &3DCNN, GRU, attention&I, PO, EVV, 2DB&attentive fusion& C, NC&PIE, JAAD\\
\hline
Abbasi \emph{et al.} \cite{DBLP:journals/corr/abs-2208-07441} &2022&arXiv& CNN, GRU&I, PO, 2DB, EVV& concat &C, NC&JAAD\\
\hline
Zhao \emph{et al.} \cite{20220811681351} &2022&IEEE SPL& Vison Transformer&I, 2DB, PO& concat &C, NC&PIE, JAAD\\
\hline
Zhang \emph{et al.} \cite{DBLP:journals/tits/ZhangAWZ22} &2022&IEEE TITS&SVM&PO, EVV, BO, W& concat & C, NC&Self-c\\
\hline
Cadena \emph{et al.} \cite{20222012127871} &2022&IEEE TITS &GCN, CNN&I, PO, 2DB, EVV, SS& concat &C, NC&PIE, JAAD\\
\hline
Zhang \emph{et al.} \cite{20223112531530} &2022&IEEE TITS &GCN& PO& concat &C, NC&JAAD\\
\hline
Yang \emph{et al.} \cite{DBLP:journals/tiv/YangZYRO22} &2022&IEEE TIV &CNN, GRU&I, 2DB, PO, EVV& attentive fusion&C, NC&JAAD\\
\hline
Achaji \emph{et al.} \cite{achaji2022attention} &2022&IV&Transformer& 2DB& concat &C, NC&PIE\\
\hline
Rasouli \emph{et al.} \cite{DBLP:conf/ivs/RasouliYR022} &2022&IV&CNN, LSTM&I, SS, 2DB, EVV&attentive fusion&C, NC&\makecell{PIE, JAAD, \\PePScenes}\\
\hline
Naik \emph{et al.} \cite{DBLP:conf/ivs/NaikBJD22} &2022&IV&GCN&I, EVV& concat & C, NC&PIE\\
\hline
Ni \emph{et al.} \cite{ni2023pedestrians} &2023&IET-ITS&CNN, GRU&I, PO& gated fusion & C, NC&PIE, JAAD\\
\hline
Ham \emph{et al.} \cite{ham2023cipf} &2023&CVPRW&CNN, GRU&I, 2DB, PO, EVV& attentive fusion& C, NC&PIE\\
\hline
Zhang \emph{et al.} \cite{DBLP:conf/aaai/ZhangTD23} &2023&AAAI&\makecell{Transformer,\\Evidential Learning}&I, 2DB& attentive fusion& C, NC&PIE, JAAD, PSI\\
\hline
Zhang \emph{et al.} \cite{DBLP:journals/corr/abs-2305-01111} &2023&arxiv&CNN, MLP&I, 2DB, PO, SS&concat& C, NC&JAAD\\
\hline
Zhang \emph{et al.} \cite{DBLP:conf/ivs/ZhangKYNMMB23} &2023&IV&CNN, MLP&2DB, O, Age, G, SS&concat& C, NC&Self-c\\
\hline
Rasouli \emph{et al.} \cite{DBLP:conf/icra/RasouliK23} &2023&ICRA&Transformer&I, SS, 2DB, T, EVV&attentive fusion& C, NC&PIE, JAAD\\
\hline
Dong \cite{dong2023pedestrian} &2023&ICLRW&Stacked GRU&I, SS, 2DB&concat& C, NC&PIE, JAAD\\
\hline
Ahmed \emph{et al.} \cite{DBLP:journals/eswa/AhmedASRH23} &2023&Expert Syst. Appl.&LSTM&I, 2DB, PO&concat& C, NC&PIE, JAAD\\
\toprule[0.8pt]
  \end{tabular}}
  \begin{tablenotes}
\item \scriptsize{
\textbf{Intention Types}: Crossing (\textbf{C}); Not-Crossing (\textbf{NC}); Walking (\textbf{W}); Standing (\textbf{ST}); Turning left (\textbf{TL}); Turning right (\textbf{TR}); Stopping (\textbf{STOP}).
\\
\textbf{Annotations:} Image (\textbf{I}); 2D Boxes (\textbf{2DB}); 3D boxes (\textbf{3DB}); Vehicle Type (\textbf{VT}); Ego Vehicle Velocity (\textbf{EVV}); Motion of Target Vehicle (\textbf{MTV}); Driver Attention (\textbf{DA}); Trajectory (\textbf{T}); Weather (\textbf{W}); Behavior (\textbf{Beh}); Pose (\textbf{PO}); Occasions (\textbf{O}); Age (\textbf{Age}); Gender (\textbf{\textbf{G}}); Depth image (\textbf{D}); Human Body Orientation (\textbf{BO}); Destination (\textbf{DES}); Semantic Segments (\textbf{SS}); Scene Text Description (\textbf{STD}).}
\end{tablenotes}
  \label{tab2}
  \end{table*}
  
\section{Agent-centric BIP}
\label{PIM-BIP}
With the background definition and key factor descriptions, this section elaborates on the progress of pedestrian-centric and vehicle-centric BIP. These two kinds of agents are studied in different observation views, different formulations, and different scenarios in this field. In particular, we present the key novelties and the latest progress.
\subsection{Pedestrian-Centric BIP}
With the successful application of deep learning, pedestrian-centric BIP methods based on Convolutional Neural Networks (CNNs), Recurrent Neural Networks (RNNs), Long Short-Term Memory (LSTM) networks, Gated Recurrent Units (GRUs), Graph Neural Networks (GCNs), and Transformer networks have become popular in this field. JAAD \cite{rasouli2017they} and PIE \cite{rasouli2019pie} datasets have the absolutely dominant position for performance evaluation. The chronological overview of the pedestrian-centric BIP methods is summarized in Table. \ref{tab2}. The categories undergo the following three stages.
\subsubsection{Spatial-Temporal Modeling in Pedestrian-Centric BIP}

To our best knowledge, Volz \emph{et al.} \cite{DBLP:conf/itsc/VolzBMGSN16} for the first time introduced Deep Neural Networks (DNNs) into this field. They employ dense neural networks for pedestrian intention classification and utilize CNNs and LSTMs to predict pedestrian crossing intention. However, the approach solely considers image inputs without incorporating other features. Compared to traditional machine learning methods such as Support Vector Machines (SVMs), their DNN-based approach achieves a 10-20\% increase in accuracy on their self-collected data. Saleh \emph{et al.} \cite{DBLP:journals/tiv/SalehHN18} transform the pedestrian-centric BIP problem into a sequential prediction task, employing stacked LSTM models to predict pedestrian crossings based on historical trajectories. Experimental results on the Daimler dataset indicate that this method exhibits lower displacement bias to the ground truth than traditional models.

Researchers have also started considering the impact of multiple sources of information on performance. Fang \emph{et al.} \cite{DBLP:conf/ivs/FangL18}, building upon image inputs, introduce the pedestrian Pose (PO) and 2D Boxes (2DB) information for pedestrian crossing intention prediction. Ablation studies reveal that the PO and 2DB information improve accuracy significantly. Furthermore, Ghori \emph{et al.} \cite{DBLP:conf/ivs/GhoriMBBDDO18} extend the work \cite{DBLP:conf/ivs/FangL18} by incorporating LSTM structures to capture the temporal dynamics of human poses. Experimental results demonstrate that the combination of spatial and temporal information achieves a 0.72 F1 score for one-second predictions on the Daimler dataset. Varytimidis \emph{et al.} \cite{DBLP:conf/sitis/VarytimidisADE18} discover that the combination of SVM and CNN is also useful for leveraging the deep features in estimating pedestrian head direction and motion and yields an 0.89 accuracy for prediction intention prediction in the JAAD dataset. Rasouli \emph{et al.} \cite{DBLP:conf/bmvc/RasouliKT19} propose a stacked GRU structure, progressively fusing pedestrian locations, ego-vehicle motion, and pedestrian appearance. Results show that the fusion of multiple sources of information, including pedestrian context, surrounding context, full context, pose, displacement, bounding box, and speed, yields very high performance (accuracy: 0.844) in the PIE dataset. Ablation experiments on feature fusion reveal that a multi-scale feature stacking in stacked GRU layers is promising in model performance. Concurrently, \cite{DBLP:conf/iccv/RasouliKKT19} demonstrate that the integration of appearance, environmental information, and pedestrian actions provides much better results in JAAD and PIE datasets. 

The aforementioned works treat the spatial and temporal by separate modules. In \cite{DBLP:conf/icra/GujjarV19,DBLP:conf/icra/SalehHN19,chaabane2020looking}, they only employ image inputs as cues but introduce 3DCNN and ConvLSTM to jointly model the spatial-temporal information for predicting pedestrian crossing intention. The experimental results generate superior performance with an average accuracy of 0.867 in the JAAD dataset to previous methods. Since then, 3DCNN and ConvLSTM became sweet pastries in other works \cite{DBLP:conf/iccvw/SinghS21,chen2021psi,20212910650116,20204009255721} in 2021. Nevertheless, because of the computation cost issue of 3DCNN, CNN, LSTM, and GRU still are the obsessive choice in \cite{DBLP:conf/ivs/RasouliYR022,DBLP:conf/ivs/ZhangKYNMMB23,dong2023pedestrian}.

\subsubsection{Interaction Modeling in Pedestrian-Centric BIP}

With the emergence of Graph Convolutional Networks (GCNs), Cadena \emph{et al.} \cite{DBLP:conf/itsc/CadenaYQW19} make early use of GCN in pedestrian crossing prediction. In their work, an adjacency matrix is computed to represent relationships between human pose key points, and these key points' coordinates are taken as input for GCN, which generates an accuracy of 0.92 in the JAAD dataset. This promotes the utility of GCN in social relationship modeling for pedestrian intention prediction and establishes a foundation for subsequent related research. Building upon \cite{DBLP:conf/itsc/CadenaYQW19,liu2020spatiotemporal,20223112531530,DBLP:conf/ivs/NaikBJD22}, GCN shows the powerful ability to model the key point relation and agent location interaction in pedestrian crossing intention prediction. Liu \emph{et al.} \cite{liu2020spatiotemporal} construct the spatiotemporal graphs with each frame's segments (different road entities) as nodes, and model the spatiotemporal relationships between pedestrians and other objects. Similarly, Zhang \emph{et al.} \cite{20223112531530} propose a human skeleton data-based spatiotemporal GCN to learn the spatial and temporal patterns simultaneously, which achieves fast inference speed and promising performance. Naik \emph{et al.} \cite{DBLP:conf/ivs/NaikBJD22}, on the other hand, employ both images and Ego Vehicle Velocity (EVV) as inputs to construct a Spatial-Temporal Scene GCN (STS-GCN) for encoding the dynamic relationships between pedestrians and other objects. Experimental results highlight the positive influence of traffic lights, traffic signs, and zebra crossings on pedestrian crossing intention. Cadena \emph{et al.} \cite{20222012127871} extend their previous work \cite{DBLP:conf/itsc/CadenaYQW19} and contribute Pedestrian-Graph+ \cite{20222012127871}. They modify the GCN structure in \cite{DBLP:conf/itsc/CadenaYQW19} to be capable of adding multimodal data information by modeling the node representation by 1D or 2D CNN for vehicle speed and image features, respectively. The ablation studies show that ego-vehicle speed affects accuracy the most. GCN models show dominant superiority in human pose and object location interaction modeling, while they cannot explicitly reduce redundant relationships. Therefore, the affinity matrix commonly involves high dimensions when the amount of objects is large with expensive computational costs.

\subsubsection{Attention Modeling in Pedestrian-Centric BIP}

Since the invention of Transformer architecture \cite{DBLP:conf/nips/VaswaniSPUJGKP17}, its capability to capture essential and attentive information has been widely explored. Pedestrian Crossing Prediction with Attention (PCPA) \cite{kotseruba2021benchmark} is a milestone work for pedestrian crossing intention prediction with attentive modeling for different information, and has become a baseline in this field. IntFormer \cite{lorenzo2021intformer} is the first work that applies the Transformer structure to pedestrian crossing intention prediction, and better performance than PCPA \cite{kotseruba2021benchmark} is obtained. The experiments show similar results of \cite{20222012127871} that self-vehicle speed is the most crucial variable for crossing intention determination. Based on this, the works of \cite{20220811681351,achaji2022attention,DBLP:conf/aaai/ZhangTD23,DBLP:conf/icra/RasouliK23} take the Transformer as the backbone model. Under this setting, the Transformer shows the promising ability for the 2D Bounding Boxes (2DB) and shows apparent improvement in the performance experiments in \cite{achaji2022attention,DBLP:conf/aaai/ZhangTD23}. In \cite{achaji2022attention}, utilizing only bounding boxes achieves an accuracy of 0.91 on the PIE dataset. Recently, the primary research team by Rasouli \emph{et al.} proposes the Pedformer \cite{DBLP:conf/icra/RasouliK23}, which combines the historical Trajectory (T), Ego Vehicle Velocity (EVV), Image (I), and Image Segments (SS) together by Multi-Head Attention (MHA) in Transformer to learn the spatial, temporal, interaction feature of pedestrians in the driving scenes. In addition, they add a trajectory prediction task to fulfill multi-task learning with pedestrian crossing intention prediction. With this formulation, Pedformer shows the highest accuracy (0.93) so far in the PIE dataset. Therefore, because of the prediction uncertainty, multi-task learning and attention modeling will become a hot pipeline in this field.

\subsection{Vehicle-Centric BIP}

 \begin{table*}[!t]\footnotesize
  \centering
  \caption{Chronological overview of the vehicle-centric BIP methods, \emph{w.r.t.}, Years, Inference Models, Multimodality Fusion Strategies, Clue Types, and Intention Types.}
  \renewcommand{\arraystretch}{1}
     \setlength{\tabcolsep}{0.01mm}{
\begin{tabular}{c|c|c|c|c|c|c|c}
\toprule[0.8pt]
Ref. &Years&booktitle&Inference Models&Clue Types&Fusion Strategies&Intention Types&Datasets\\
\hline
Dang \emph{et al.} \cite{dang2017time} &2017&ITSC &LSTM&EVV, HD& concat &LLC, RLC&Self-c\\
\hline
Hu \emph{et al.} \cite{DBLP:conf/ivs/HuZT18} &2018&IV &CNN, GMM&EVV, D2C& concat &LLC, RLC, LK&NGSIM\\
\hline
Casas \emph{et al.} \cite{DBLP:conf/corl/CasasLU18} &2018&CoRL& CNN& 3DP, HD& concat &\makecell{LK, TR, TL, LLC, \\RLC, STOP, P}&Self-c\\
\hline
Tang \emph{et al.} \cite{tang2018lane} &2018&Expert Syst. Appl.&\makecell{Adaptive Fuzzy \\Neural Network}&EVV, D2V, SA& concat &LLC, RLC&Self-c\\
\hline
Scheel \emph{et al.} \cite{DBLP:conf/icra/ScheelSNT18} &2018&ICRA&Bi-LSTM& EVV, LoC& one-hot vector &LLC, RLC&NGSIM\\
\hline
Tang \emph{et al.} \cite{DBLP:journals/eswa/TangYLCH19} &2019&Expert Syst. Appl.&MLP+Fuzzy C-Means&EVV, D2V, SA& concat &LLC, RLC&Self-c\\
\hline
Han \emph{et al.} \cite{han2019driving} &2019&IV&LSTM&T, EVV& concat &LLC, RLC&NGSIM\\
\hline
Izquierdo \emph{et al.} \cite{DBLP:conf/itsc/IzquierdoQPLS19a} &2019&ITSC&CNN, LSTM& T& muti-Channel Stacking &LLC, RLC&PREVENTION\\
\hline
Zyner \emph{et al.} \cite{DBLP:journals/tits/ZynerWN20} &2020&IEEE TITS&RNN& T, EVV, SA& concat &LK, TL, TR, UT&Self-c\\
\hline
Mahajan \emph{et al.} \cite{mahajan2020prediction} &2020&Transport. RR&LSTM& T, EVV, D2C& concat &LK, TL, TR, UT&HighD\\
\hline
Girma \emph{et al.} \cite{DBLP:conf/ivs/GirmaAWKH20} &2020&IV& LSTM +Attention&EVV& Muti-Channel Stacking  &LK, TL, TR, STOP&NDS\\
\hline
Li \emph{et al.} \cite{DBLP:journals/tvt/LiZXWCD21} &2021&IEEE TVT &RNN &RL, EVV, SA& concat & LLC, RLC&Self-c\\
\hline
Griesbach \emph{et al.} \cite{DBLP:journals/tits/GriesbachBH22} &2022&IEEE TITS&Echo State Network, LSTM& SA& concat & LLC, RLC&Self-c\\
\hline
Chen \emph{et al.} \cite{chen2022intention} &2022&IEEE TITS&LSTM+Attention&T, EVV, VT& concat &LLC, RLC&NGSIM, HighD\\
\hline
Jiang \emph{et al.} \cite{jiang2023intention} &2022&TRR&Transformer&T& concat &LC, ACCE, DECE&NGSIM, HighD\\
\hline
Hu \emph{et al.} \cite{hu2022causal} &2022&ICRA &Variational RNN&HD, EVV& causal &MRD, Y, M&INTERACTION\\
\hline
Wang \emph{et al.} \cite{DBLP:journals/tie/WangQYLXH22} &2022&TIE &LSTM&T,EVV& concat &LLC, RLC&NGSIM\\
\hline
Gao \emph{et al.} \cite{DBLP:journals/tits/GaoLCHLDL23} &2023&IEEE TITS&Transformer&EVV, D2V, T& concat &LLC, RLC&NGSIM, HighD\\
\hline
Lu \emph{et al.} \cite{DBLP:journals/tits/LuHYS23} &2023&IEEE TITS&SVM&EVV, LoC& concat &CI, CO&Self-c\\
\hline
Li \emph{et al.} \cite{DBLP:journals/tiv/LiWZ23} &2023&IEEE TIV&Transformer&EVV, T, MTV&attentive fusion&CI, CO&NGSIM, HighD\\
\hline
Do \emph{et al.} \cite{do2023lane} &2023&IEEE TIV&LRLSE*&EVV, T, MTV&probability fusion&LLC, RLC&HighD\\
\toprule[0.8pt]
  \end{tabular}}
  \begin{tablenotes}
  \scriptsize{
\item \textbf{Intention Types}: Lane Keeping (\textbf{LK}); Turning Left (\textbf{TL}); Turning Right (\textbf{TR}); U-Turn (\textbf{UT}); Cutting In (\textbf{CI}); Cutting Out (\textbf{CO}); Vehicle Overtaking (\textbf{VO}); Left Lane Changing (\textbf{LLC}); Right Lane Changing (\textbf{RLC}); Stopping (\textbf{STOP}); Pushing (\textbf{P}); Yielding (\textbf{Y}); Merging (\textbf{M}); Moving along the Roundabout (\textbf{MRD}); Accelerating (\textbf{ACCE}); Decelerating (\textbf{DECE}).\\
\item \textbf{Clue Types}: Ego Vehicle Velocity (\textbf{EVV}); Motion of Target Vehicle (\textbf{MTV}); Trajectory (\textbf{T}); Distance to Centerline (\textbf{D2C}); Distance to Vehicles (\textbf{D2V}); 3D Point Cloud (\textbf{3DP}); Road Map (\textbf{HD}); Steering Angle (\textbf{SA}); Road Lines (\textbf{RL}); Longitudinal Coordinate (\textbf{LoC}); Vehicle Type (\textbf{VT}).\\
LRLSE*: Linearized recursive least square estimation.}
\end{tablenotes}
  \label{tab3}
  \end{table*}
  
Unlike the pedestrian-centric BIP task, vehicle-centric BIP primarily focuses on intention types such as Lane Changing (LC), Merging (M), Turning Left/Right (TL/TR), and Lane Keeping (LK). Table. \ref{tab3} presents a chronological overview of the vehicle-centric BIP methods.

\subsubsection{Surrounding Vehicle (SV)-Centric BIP}
\begin{figure}[!t]
 \centering
 \includegraphics[width=\hsize]{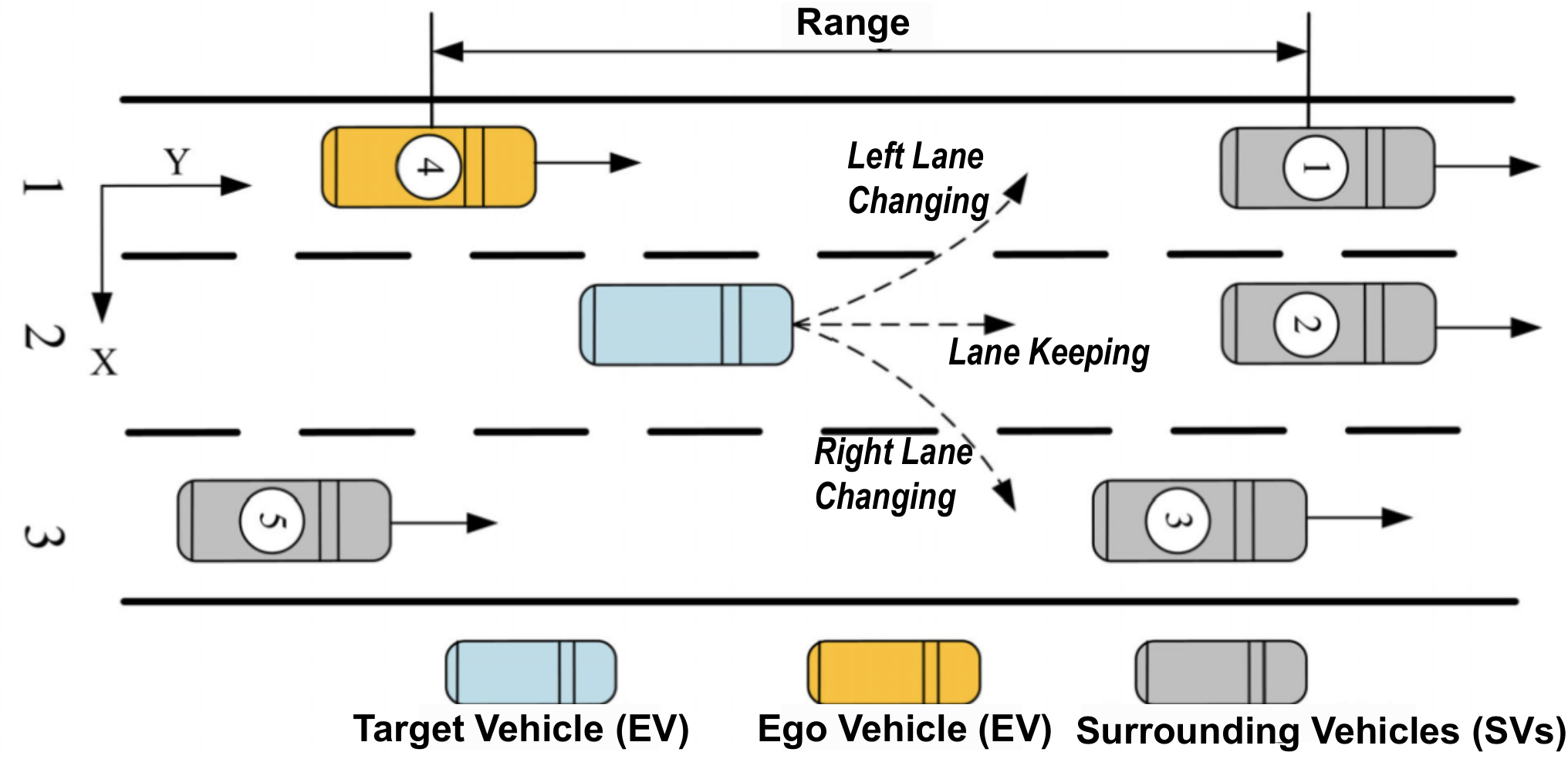}
  \caption{The illustration for the Lane Changing (LC) and Lane Keeping (LK) intentions of the Target Vehicle (TV), where the Ego Vehicle's (EV) movement is strongly influenced by the behavioral intention of TV. This figure is credited to \cite{DBLP:journals/tits/GaoLCHLDL23}. }
  \label{fig10}
\end{figure}

Surrounding Vehicle (SV)-centric BIP can provide an interactive understanding of scenes for the Ego Vehicle (EV) with a BEV observation, as shown in Fig. \ref{fig10}, where the NGSIM and HighD are two common datasets in these situations. In particular, the Lane Changing (LC) intention and Lane Keeping intention are two primary types. Because of the trajectory data form in this situation,  sequential networks, such as LSTM, are popular for modeling the temporal locations of vehicles \cite{DBLP:journals/tvt/LiZXWCD21}. For example, Dang \emph{et al.} \cite{dang2017time} treat LC prediction as a regression problem, which employs an LSTM network to predict the Time-To-Lane-Change (TTLC) by incorporating driver status, vehicle information, and environmental cues. Through the ablation experiments, this work obtains a 3.2\% improvement in F1 score over the traditional SVM method in a self-collected dataset. Scheel \emph{et al.} \cite{DBLP:conf/icra/ScheelSNT18} introduce the Bidirectional LSTM (BiLSTM) to make a cross-check for temporal correlation modeling the relation between vehicle trajectory points and LC intention. Compared with LSTM, BiLSTM obtains 4\%improvement (achieving an accuracy value of 0.926) on the NGSIM dataset. Furthermore, the interaction between vehicles is considered in \cite{han2019driving}, which infers the Hybrid State System (HSS) between the Target Vehicle (TV) and SVs. Based on the experiments on the NGSIM dataset, the accuracy for LLC, RLS, and LK intention reaches 0.0.953, 0.980, and 0.963, respectively. Similarly, the social interaction of SVs is also modeled in recent works \cite{chen2022intention,jiang2023intention,DBLP:journals/tiv/LiWZ23}, while differently, they take the Transformer to fulfill an attentive feature extraction for maneuver prediction and maneuver-aware trajectory prediction of SVs. The ablation results on the NGSIM and HighD datasets demonstrate that the interaction-inclusive module promotes the BIP significantly.

SV-centric BIP commonly takes the trajectory under BEV observation as input, the LC, LK, CO, and CI intentions can be directly measured without geometrical distortion. However, the Surrounding-Vehicle (SV) intention is based on the exhibited behaviors, \emph{i.e.}, that the SV-centric BIP is more like behavior prediction with a short prediction horizon. Contrarily, the Ego-Vehicle (EV)-centric BIP can leverage the maneuver status of the vehicle itself to fulfill more reasonable BIP.
\subsubsection{Ego Vehicle (EV)-Centric BIP}
For the Ego Vehicle (EV)-centric BIP, different from SV-centric BIP, the steering angle is a focused intention indicator. For example, an Adaptive Fuzzy Neural Network (AFFN) \cite{tang2018lane} fuses vehicle sensor data to predict Steering Angles (SA) of the Ego Vehicle (EV), thereby achieving LC intention prediction. In addition, compared with the monotonous LC or LK intention types, EV-centric BIP involves more diverse intention types. In this category, some formulations utilize voxelized 3D LiDAR data to obtain a rasterized road map. The Intent Network (IntentNet) \cite{DBLP:conf/corl/CasasLU18} is a typical method that predicts seven kinds of intentions of EV by inputting BEV representation from the voxelized LiDAR data and the High-Definition (HD) map. Ablation experiment results demonstrate that the HD map can significantly enhance model performance, which verifies the crucial role of road structure representation in vehicle-centric BIP. Furthermore, EV-centric BIP can encode the camera videos, road maps, road entity segments, and social interaction in model inference. The scene context feature can be obtained to facilitate the accurate BIP. For example, Izquierdo \emph{et al.} \cite{DBLP:conf/itsc/IzquierdoQPLS19a} introduce a CNN-LSTM model to encode RGB video frames, local and global scene context features, and temporal information of agent to achieve LC intention prediction, where the PREVENTION dataset provides a Cutting In (CI) and Cutting Out (CO)  intention labels. 
 \begin{figure*}[!t]
  \centering
 \includegraphics[width=\hsize]{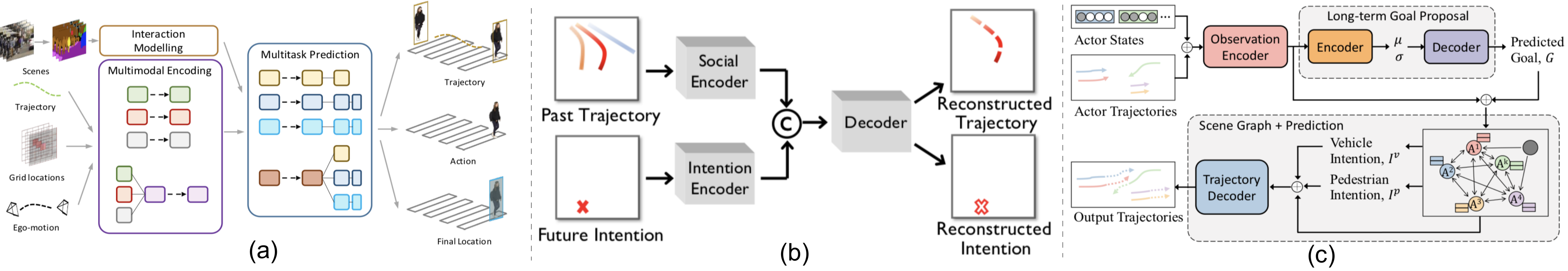}
  \caption{Multi-task learning prototype for BIP-Conditioned Trajectory Prediction, where (a) is the BiPed for simultaneous prediction of pedestrian trajectory, intention, and final location in the Ego-View \cite{DBLP:conf/iccv/RasouliRL21}, (b) is the joint reconstruction of future trajectory and future intention points in \cite{DBLP:journals/corr/abs-2203-11474} in BEV observation. (c) is a goal intention conditioned trajectory prediction framework in \cite{DBLP:conf/iccv/GiraseGM0KMC21}.}
  \label{fig11}
\end{figure*}

\subsection{In Summary}

To enhance the category margin of different intentions, most existing works exploit multiple clues. As for pedestrians and vehicles, the types of clues are different.  The inference of pedestrian intention prediction is prone to use images (I), pose (P), and 2D Boxes (2DB), while vehicles often fuse many road structure information, such as images (I), HD map (HD), Distance to Centerline (D2C), and vehicle velocity (\emph{i.e.}, EVV or MTV). From Table. \ref{tab2} and Table. \ref{tab3}, it is apparent that the intention types of pedestrians and vehicles are different, where Crossing (C) and Not Crossing (NC) are the main concerns of pedestrians, but Lane Changing (LC) is the target of interest for vehicles. This observation is reasonable, where the``\emph{crossing warning}" is a necessary function for assisted driving systems \cite{kwak2017pedestrian}, and lane changing of vehicles is the most frequent behavior with the potential threat to the other vehicles \cite{DBLP:journals/tvt/XingLWWACVW19}. 

Most multi-clue intention prediction works do not evaluate the importance of different clues but simply fuse them in a ``\emph{concat}" strategy. Contrarily, ``\emph{attentive fusion}" provides a mechanism for selecting the important information, and performs better adaptation for different situations. 

In retrospect, Deep Neural Networks (DNNs) have been widely applied in agent-centric BIP. Although various neural network models have emerged, those CNNs, LSTMs, and GRUs continue to demonstrate impressive performance. The diverse input sources and feature fusion strategies have proven critical to model performance. Attentive fusion methods have shown notable enhancements and have attracted repeated concentration in recent works. The utilization of widely adopted benchmarks such as JAAD and PIE has shifted the evaluation of models from self-collected datasets to a more consistent and comparable framework.

\section{BIP-Aware Applications}
\label{tspre}
The accurate BIP provides the future movement tendency of road agents. It is useful for the following trajectory prediction and behavior prediction tasks. How BIP promotes the other prediction tasks will be described here.
\subsection{BIP-aware Agent Trajectory Prediction}
The first prediction task is BIP-aware trajectory prediction, where the problem can be currently inferred by a multi-task learning prototype and a parameter conditioning prototype.
\subsubsection{Multi-task Learning Prototype}
Formulating the joint prediction of behavioral intention and trajectory as a multi-task learning prototype can be easily considered and implemented. By adding an extra loss function with trajectory prediction loss, these two coupling tasks can be inferred simultaneously \cite{tian2022multi,su2022crossmodal,choi2020drogon,sui2021joint,DBLP:journals/corr/abs-2010-10270}. 

In the Ego-View situations, the pedestrian crossing intention prediction recently leverages the future location or trajectory to make performance assistance. For example, Su \emph{et al.} \cite{su2022crossmodal} treat the pedestrian crossing intention as an extra signal and fulfill the trajectory prediction by adding an intention loss (cross-entropy of the intention labels) to the endpoint with L2 loss of trajectory. The results show that crossing intention has promoted the trajectory prediction for the end-time step significantly. Rasouli \emph{et al.} \cite{DBLP:conf/iccv/RasouliRL21} formulate a multi-task prediction (BiPed) for pedestrian crossing intention, trajectories, and final grid location, as shown in Fig. \ref{fig11}(a). BiPed improves the performance of pedestrian crossing intention prediction and trajectory prediction together. The binary cross-entropy loss is used for the pedestrian crossing intention prediction. Sui \emph{et al.} \cite{sui2021joint} introduce the Transformer to model the cross-attention of different information (locations and images) and also formulate the multi-task learning of pedestrian crossing intention and trajectory prediction. PedFormer \cite{DBLP:conf/icra/RasouliK23} is a new work for pedestrian motion prediction with multi-task learning, where the learned feature is decoded by a Hybrid Gated Decoder constructed by stacked LSTM for crossing intention and future trajectory prediction. 

Under the BEV observation, one kind of formulation for BIP-conditioned trajectory prediction is to leverage the future intended goal points to guide the prediction. The Retrospective-Memory-based Trajectory Prediction (RememNet) \cite{DBLP:journals/corr/abs-2203-11474} combines the future intended goal points (named as \emph{future location intention}) and trajectory prediction together, and infers the intention prediction with the MemoNet to reconstruct the compatible future trajectory and future intended goal points jointly, as shown in Fig. \ref{fig11} (b). The results of RememNet demonstrate that a suitable number selection of intended goal points is important for avoiding the intention modality missing and irrelevant instances. DROGON \cite{choi2020drogon} fulfills a goal-oriented trajectory prediction network, which computes the probability of intended goal points based on the inferred interaction of vehicles, and estimates the label of future intention goal points by cross-entropy loss. 
Actually, the aforementioned goal-oriented trajectory prediction is also another kind of intention for conditioning the trajectory prediction. The goal estimation is also fulfilled by the crossing-entropy loss \cite{DBLP:conf/corl/ZhangSHHM20,DBLP:conf/iccv/Gu0Z21,DBLP:conf/corl/ZhaoGL0SVSSCSLA20}, as investigated in Sec. \ref{goal}. 

\subsubsection{Parameter Conditioning Prototype}
Parameter Conditioning Prototype for trajectory prediction usually models the intent as extra information to re-weight or re-constrain the trajectory distribution sampling function \cite{li2021grin,liu2019integrated,huang2020long}. The Conditional VAE (CVAE) models \cite{lopez2020decision} defined as follows are commonly adopted.
\begin{equation}
p_\theta(y_i|{\bf{X}})=\int p_\theta(y_i|{\bf{z}}_i,{\bf{X}})p_\theta({\bf{z}}_i|{\bf{x}}_i)d{\bf{z}}_i,
\end{equation}
where $p_\theta({\bf{z}}_i|{\bf{x}}_i)$ denotes the conditional independence of the latent variables ${\bf{z}}_i$ under the agent observation ${\bf{x}}_i\in {\bf{X}}$. Commonly, the intention is encoded in $p_\theta({\bf{z}}_i |{\bf{x}}_i)$, where the other conditions, such as interaction and road scene knowledge may also be encoded. 
Euro-PVI \cite{DBLP:conf/cvpr/BhattacharyyaRF21} models the interactive intention between the surrounding objects and ego-vehicles (\emph{e.g.}, \emph{yielding, decelerating, and crossing, etc}.), and develops a Joint-$\beta$-CVAE to conduct the trajectory prediction, where the interaction intention is encoded as the latent variables in the CVAE formulation. The results verify that involving the interactive intention between pedestrians and vehicles could significantly reduce the ADE and FDE values. Sun \emph{et al.} \cite{sun2022domain} also propose a CVAE model to jointly predict the intended goals and trajectories, which embeds the predicted goals and the interaction of agents with a Multiple layers Perception (MLP) at each time step. Recent work LOKI \cite{DBLP:conf/iccv/GiraseGM0KMC21} treats the intended goals as a condition for scene graph construction, where the outputs of a Goal Proposal Network (GPN) and the agent intention prediction model are added to decode future trajectories, as shown in Fig. \ref{fig11}(c).

Recently, the trajectory prediction of vehicles in highway scenarios usually takes the parameter conditioning prototype. For example, Gao \emph{et al.} \cite{DBLP:journals/tits/GaoLCHLDL23} propose a dual Transformer to encode the past trajectories and interactive information, and decode the future Lane Changing (LC) intention. Then, the predicted LC intention probability is fed into the feature of past trajectories for the subsequent trajectory prediction. Compared with the naive Transformer, the Root-Mean-Squared Error (RMSE) value for future predicted trajectories is reduced by 7.52\% and 27.3\% on the NGSIM and HighD datasets, respectively \cite{DBLP:journals/tits/GaoLCHLDL23}. Do \emph{et al.} \cite{do2023lane} take the LC intention as a prior for future trajectory prediction. Differently, they initialize the path generation step by a cubic spline curve in the Frenet Coordinate System (FCS), then predict the LC intention and future trajectories by a dynamic estimation of path probabilities.

Besides, other parameter-based intention prediction models, such as the Dynamic Bayesian Network (DBN)~\cite{schulz2018multiple,wu2020crossing,hu2022causal,tang2015turn,xu2021roadside}, are also explored. As for the deep learning era, the framework of DBN will be popular, where the feature extraction of the inference model may be fulfilled by deep learning modules. The work \cite{shen2022parkpredict+} firstly predicts the vehicle intention on the BEV sequence by a CNN model with the binary cross-entropy loss, and then fuses the predicted intent to the trajectory prediction with a multi-head attention decoder model. Ma \emph{et al.} \cite{ma2021continual} propose a continual multi-agent behavior prediction work, which designs an episodic memory buffer and a conditionally generative memory to capture the historical interaction trajectories with the labeling of goal position and interaction intention. Wu \emph{et al.} \cite{wu2020crossing} fuse the pedestrians' behavior, intention, and scene context together to tackle the trajectory prediction problem. The pedestrian intention is inferred by DBN with the variables for the existence of the crossing area, waiting time, distance to curb, \emph{etc.} The pedestrian crossing intention is treated as a bool variable to change the trajectory sampling function. In some works, the researchers fuse the intention and trajectory prediction as a sequential prediction problem, where the predicted trajectories are also useful for the intention prediction tasks. For example, Saleh \emph{et al.} \cite{DBLP:journals/tiv/SalehHN18,saleh2017intent} predict the long-term intention of pedestrians by a stacked LSTM over the trajectory points. 

\textbf{Discussion:} Multi-task learning and parameter conditioning prototypes in BIP-aware trajectory prediction become popular recently. From the task jointing strategy for BIP-aware trajectory prediction, the cross-entropy loss is commonly utilized to optimize the BIP task. In the multi-task learning prototype, the parameters of modules on two tasks commonly share the weights. Contrarily, the parameter conditioning prototypes usually have separate weight sets. Explainability still is a core issue in the multi-task learning prototype. The parameter conditioning prototype seems to be appropriate because of the flexible dynamic networks and can provide a parameterized explainability for information importance modeling. Certainly, the parameter conditioning prototype can also involve multi-task learning to optimize the models. 
\subsection{BIP-aware Agent Behavior Prediction}

In many related works, the terms behavior prediction and trajectory prediction have been interchanged and used by treating trajectory prediction as behavior prediction. We think these two tasks have intrinsic differences, where behavior prediction is a classification problem but trajectory prediction is commonly a regression problem. 

  \begin{figure}[!t]
  \centering
 \includegraphics[width=\hsize]{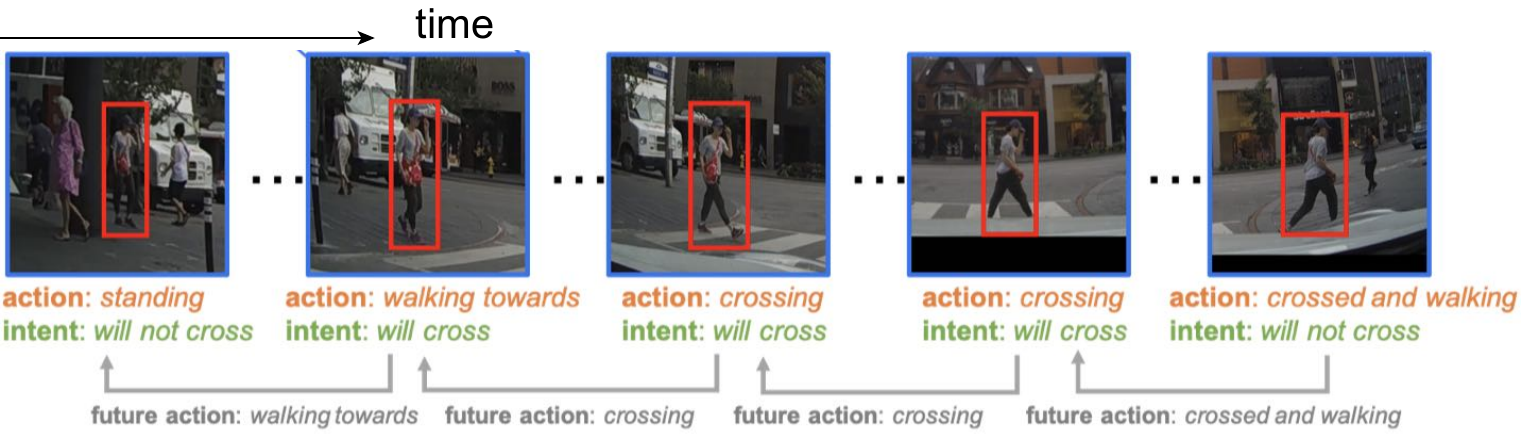}
  \caption{Crossing intention aware pedestrian behavior prediction. The crossing intention is coupled with crossing behavior \cite{DBLP:conf/ijcai/YaoAJV021}.}
  \label{fig12}
\end{figure}
Compared with trajectory prediction, behavior prediction has the most similar problem formulation with BIP, while behavior prediction determines the behavior label within a longer time window \cite{DBLP:journals/tits/Li0YG22,DBLP:conf/ivs/HuZST19}. Besides, since the behavior may last for a while, the behavior prediction can also be formulated as a sequential classification task for several future time steps, \emph{e.g.}, that the prediction changes from the ``\emph{will cross}" to ``\emph{crossing}" for pedestrians. As shown in Fig. \ref{fig12}, Yao \emph{et al.} \cite{DBLP:conf/ijcai/YaoAJV021} couple the crossing \emph{intention} and crossing \emph{behavior} of pedestrians, where the \emph{``standing", ``walking towards", ``crossing"} and \emph{``crossed"} actions are combined with the intention of ``will cross or not cross". The pedestrian behavior prediction is modeled as a sequential prediction problem solved by a multi-task inference and verifies that intention affects actions and future action is also useful for accurate intention prediction. Inspired by \cite{DBLP:conf/ijcai/YaoAJV021}, Zhai \emph{et al.} \cite{zhai2022social} also formulate a multi-task learning model for pedestrian crossing intention and behavior prediction. Differently, they propose a Spatial-Temporal Heterogeneous Graph (STHG) to model the relationships between pedestrians and surrounding dynamic and static road entities and improve the Average Precision (mAP) of \cite{DBLP:conf/ijcai/YaoAJV021} to 0.26 (+0.03) on the JAAD dataset for pedestrian behavior prediction within one second time.  
Banijamali \emph{et al.} \cite{banijamali2021prediction} construct an action-conditioned behavior prediction, where the prediction problem is formulated as latent probabilistic generative process $p({\bf{o}}_{t+1}|{\bf{o}}_{1:t}, {\bf{a}}_t)$ with the action ${\bf{a}}_t$ and observation feature ${\bf{o}}_{1:t}$. The action ${\bf{a}}_t$ at time $t$ and the future state ${\bf{o}}_{t+1}$ are alternatively predicted to fulfill a ``Prediction by Anticipation" framework.  Li \emph{et al.} \cite{li2020behavior} propose an interaction and behavior-aware driving behavior prediction framework based on joint predictions of intentions and motions of surrounding vehicles, which is fulfilled by a multi-modal hierarchical Inverse Reinforcement Learning (IRL) over the driving trajectory data. The driving behaviors are defined as aggressive, conservative, and moderate driving.

From the investigation, we find that compared with BIP-aware trajectory prediction, the research on BIP-aware behavior prediction is limited. Sometimes, this field treats behavioral intention and behavior prediction in the same concept, while they are rather different from the problem connotation \cite{ajzen1974factors}. Manifestly, the behavioral intention of road agents has a positive promotion role for long-term behavior prediction. In addition, similar to BIP-aware trajectory prediction, current BIP-aware behavior prediction works also utilize multi-task learning or parameter-conditioning prototypes. Differently, behavior prediction is a classification problem, where the explainability of models, such as the explanation of the importance of spatial-temporal regions, will be promising for trustworthy prediction. However, this issue is not explored in current works.

\section{Extension and Discussion}
\label{fcon}
Through the exhaustive investigation of behavioral intention prediction and its roles in other prediction tasks, we arrive at a full portrait of this topic. Here, we make an extension and discussion for BIP.
\subsection{Benchmarks and Theories}
\subsubsection{Benchmarks}
As discussed in Sec. \ref{datasets}, most of the available benchmarks for BIP focus on the behavioral intention of Crossing (C), Not Crossing (NC) for pedestrians, Lane Changing (LC), and Lane Keeping (LK) for vehicles. In addition, the data observation views concentrate on the Ego-View, which cannot capture the full range of the road scene, and many types of behavioral intentions cannot be found, such as the ``rear car following", ``overtaking from behind", \emph{etc.} A possible way is to add the behavioral intention label for the datasets with panoramic views, such as the Argoverse 3D dataset \cite{chang2019argoverse}, nuScenes dataset \cite{caesar2020nuscenes}, or KITTI-360 \cite{liao2022kitti}. It is also interesting to introduce viewpoints from novel devices, \emph{e.g.}, drone and satellite, for comprehensive scene understanding \cite{zheng2020university}.  Besides, the intention types in current datasets are not fine-grained enough, and the intention type imbalance issue is universal. In the future, more fine-grained interactive intention types between road agents with other road entities can be considered. For example, pedestrians of different ages and genders often show different behavioral intentions on the road. Furthermore, the safe-critical scenarios with long-tailed distribution or harsh environments (\emph{e.g.}, rainy, foggy, snowy, windy, and low-light conditions) also need to be considered.

In addition, It is noted that all current works evaluate the performance on different datasets, and the evaluations have obvious performance gaps because of data shifts \cite{kotseruba2021benchmark}. 
For cross-dataset evaluation on pedestrian crossing intention prediction, a recent work \cite{DBLP:conf/ivs/GesnouinPSM22} shows that current state-of-the-art pedestrian crossing prediction models generated poor performance in cross-dataset evaluation (JAAD and PIE). They introduce the confidence calibration metrics, \emph{i.e.}, Expected Calibration Error (ECE) and Maximum Calibration Error (MCE) \cite{DBLP:conf/aaai/NaeiniCH15}, to provide a complement evaluation, and find that ECE and MCE differ drastically. In the meantime, the pre-trained model on diverse source datasets can boost the generalization ability of target datasets. For the BIP problem, besides the crossing intention types, multiple kinds of behavioral intentions and the uncertainty estimation of model calibration in multi-label classification problems need to be explored. Furthermore, with the development of deep learning models, the data-model calibration measurement \cite{DBLP:conf/cvpr/NixonDZJT19} is also a core issue in trustworthy implementation.
\subsubsection{Theories}
Despite the numerous works on BIP that have exhibited significant progress in performance, most of the current works on BIP are all based on CNN, LSTM, Conv-LSTM, Transformer, GCN, \emph{etc.}, as described in Sec. \ref{PIM-BIP}. These deep learning models are all deterministic neural networks for achieving a mapping from input space to output space, which is usually overconfident in the testing phase. Consequently, one self-calibrated deep learning approach on one benchmark faces the data shift issue in the evaluation and may cause over-fitting or under-fitting problems when encountering the dataset with simpler and more diverse samples, respectively. 

For the deterministic neural networks, current research efforts employ domain adaptation to address this problem by using a well-pre-trained model on large-scale datasets or leveraging more complex architectures. For example, vision-language pre-trained models, such as BEit-3 \cite{wang2022image} and VinVL \cite{zhang2021vinvl}, learn an informative representation with the help of dense semantics in language. However, although these pre-trained models can generate a good representation, the domain gap in BIP is still large and needs further valuable inference models. We think the possible ways for developing the new theories on BIP should consider the influencing factors (described in Sec. \ref{ifbi}) as aforementioned, such as the better adoption of the road structure representation, social interaction modeling, and robust estimation of the prediction uncertainty. Standing at the natural characteristics of multiple clues and preferring aims in the BIP problem, more explainable scene representation with scene knowledge can be involved, such as the scene graph \cite{DBLP:conf/eccv/DevaranjanKF20}. 

Essentially, fusing more clues could reduce the aleatoric uncertainty as aforementioned. More information provides more constraints for future intention prediction, while it gives rise to a fundamental problem on how to fuse this information in the best way. That is because, in some situations, some information may be counteractive. Various Dynamic Neural Networks (DNNs) \cite{han2021dynamic} may be promising for adaptively selecting multi-modal information in different situations. Dynamic Multimodal Fusion (DynMM) \cite{xue2022dynamic} and Dynamic Routing Network (DRN) \cite{cai2021dynamic} are two kinds of models, where the ``dynamics" in modality fusion is fulfilled by a Gating Network (GN) in DynMM and achieved by the router network in DRN. The gating network will select the best expert network in the final decision. Certainly, the gating network can also be added in the feature embedding part to fulfill a selective multi-modal encoding.

Besides the deterministic neural networks, stochastic neural networks aim to estimate the prediction distribution, which is possible for providing solutions for the prediction uncertainty estimation, caused by data shift, Out-of-Distribution (OOD) samples (\emph{i.e.}, unfamiliar behavioral intention), the objective property of behavioral intention, and the long-term prediction situations. Existing works \cite{zhang2021dense} estimate the distribution uncertainty for the Bayesian neural networks, generative adversarial networks, CVAE, or deep ensembles \cite{DBLP:conf/nips/Lakshminarayanan17} by adding the uncertainty consistency loss in the Bayesian latent variable model \cite{depeweg2018decomposition}. Therefore, a possible direction is to develop the models with more consideration of prediction uncertainty.

\subsection{Parallel Testing}
\label{rssimulation}
Parallel testing refers to a real-synthetic data collaboration for BIP formulation and evaluation. As aforementioned in Sec. \ref{ifbi}, we need to find the natural relation of the road entities and collect sufficient data samples. However, in practical use, it is difficult to gather adequate samples that cover all of the causal relation, diversity, and long-tailed behavioral intention types in safety-critical driving scenes. Consequently, the data imbalance issue is haunting us all the time. Therefore, more and more works begin to borrow multitudinous virtual simulation tools (\emph{e.g.}, CARLA \cite{DBLP:conf/corl/DosovitskiyRCLK17}, GTA-V \cite{DBLP:conf/iccv/RichterHK17}, \emph{etc.}) to generate diverse driving scenes in this field. We call it Simulation Augmentation (SA) in this paper. 

The core problems in SA are to transfer the scene consistency from real to synthetic data, and transfer the diversity from synthetic to real scenarios, which generates the possible future state with a parallel evolution \cite{DBLP:journals/tiv/Wang0SW0T00XW0022}. 

\subsubsection{Real-to-Synthetic Generation}
Within this domain, many kinds of virtual engines are adopted with high-fidelity rendering \cite{DBLP:journals/tits/DingXAL0Z23}.
With these excellent simulators in the driving scene, the research on BIP may be promoted significantly in future years. For example, Chen and Krahenbuhl \cite{chen2022learning} create a virtual multi-vehicle collaboration environment for the BIP of Ego Vehicle (EV) with the learning of the future intention of Surrounding Vehicles (SVs). TrafficSim \cite{suo2021trafficsim} can flexibly generate the behaviors of \emph{``U-Turn", ``Yielding", and ``Merging", etc.}, for road vehicles. 

SA is gradually becoming an indispensable technique for the reasoning of safe-critical driving scenarios. It has a direct relation with Digital Twining (DT) \cite{xie2021digital} or Parallel Intelligence (PI) \cite{wang2016steps} in the driving scene. With the booming of the Metaverse, the interaction between the virtual and real world will become a core basis for understanding the world.

\subsubsection{Synthetic-to-Real Adaptation}
Synthetic-to-real adaptation can absorb the superiority of various simulators for generating vast amounts of data in different weather, light, and road conditions. The data with long-tailed distribution or adverse weather conditions can be collected efficiently. 

Actually, this field has made some attempts at pedestrian crossing intention prediction with the assistance of synthetic data. For example, the work \cite{achaji2022attention} transfers the dynamics of the bounding box from synthetic data to real data. Another work \cite{baijie2022} constructs 4667 sequences with ``C" or ``NC" intention and models a virtual-to-real deep distillation for the lightweight pedestrian crossing intention prediction. Different from the works that address the BIP with the Ego-View observation, Kim \emph{et al.} \cite{kim2020pedestrian} propose a pedestrian crossing intention prediction model with the pedestrians' view with a Virtual Reality (VR) apparatus. Although synthetic data can boost the diversity of the scenarios, there is a large distribution gap between the synthetic data and real data. Therefore, models trained on synthetic data often show degraded generalization to real data \cite{DBLP:conf/iclr/ChenYMLAWA21}. Recently, Zhou \emph{et al.} \cite{zhou2022domain} present a survey for the domain generalization problem and exhibit the core solutions for better synthetic-to-real adaptation.

\subsection{Counterfactual Analysis}
\subsubsection{Causality Inference in BIP}
From the investigation of the agent-centric BIP works in Sec. \ref{PIM-BIP}, we can see that there are few attempts to consider the explainable models. The attention mechanism (\emph{e.g.}, self-attention) shows an initial beginning for important feature learning \cite{zablocki2022explainability}. However, what clue is crucial for BIP? Causality inference may be a promising choice.

\begin{figure}[!t]
  \centering
 \includegraphics[width=\hsize]{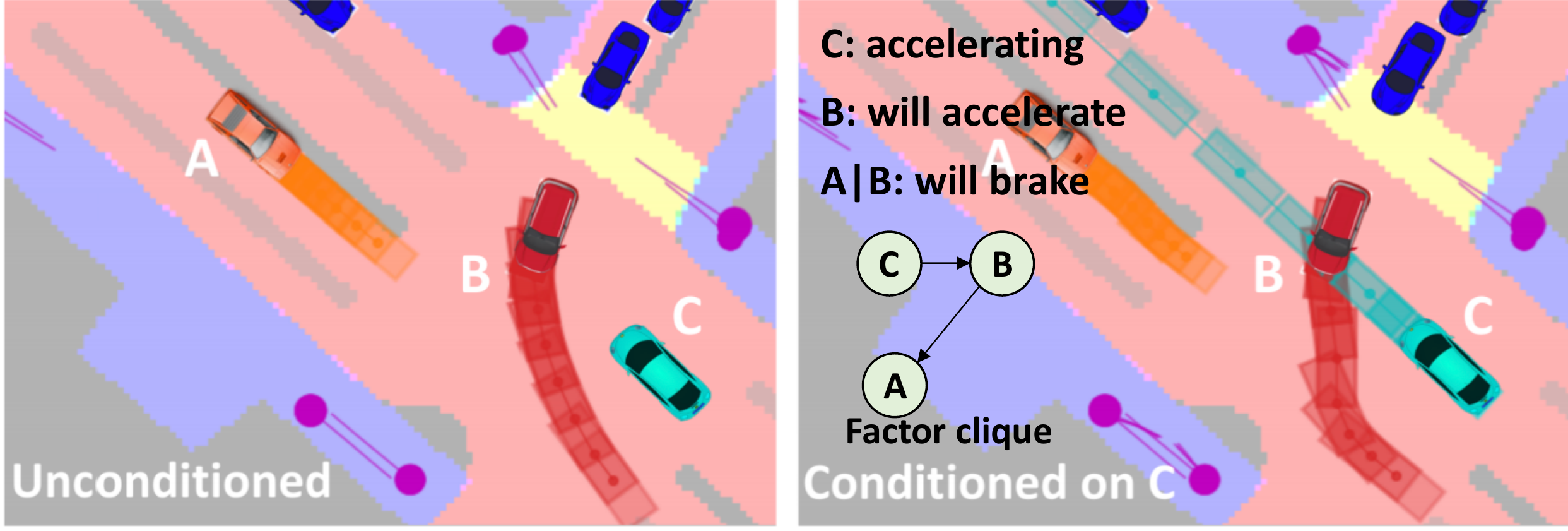}
  \caption{The conditioning analysis for future prediction with a causal intention linking to different vehicles. Credits to \cite{moraffah2020causal}.}
  \label{fig13}
\end{figure}

Causal \cite{hu2022causal} or factor relation \cite{chen2022scept} amongst the road agents is involved in future state prediction, and constructs the non-visible ``Dark Matter" \cite{DBLP:journals/pami/XieSTZ18} for motivating the behavioral intentions. Chen \emph{et al.} propose scene-consistent, policy-based trajectory predictions, which firstly build a scene graph by the agents' distance, and partition the graph into several cliques. The factor graph is constructed on these cliques, which paved the way for the \emph{conditioning} and \emph{counterfactual} analysis \cite{moraffah2020causal} for the prediction. Taking Fig. \ref{fig13} as an example, the intention of the vehicle \textbf{A} will be influenced by the causal chain of the vehicle \textbf{B} conditioned by the accelerating vehicle \textbf{C}. Hu \emph{et al.} \cite{hu2022causal} contribute a causal-based time series domain generalization model for vehicle intention prediction. The causal knowledge originates from the road topology, speed limit, and traffic rules. 

With the causal or factor relation, \emph{counterfactual} analysis can find the primary cause or the scene knowledge for the specific prediction results by imagining a change in the input state. For example, Li \emph{et al.} \cite{li2020make} explore the causality on the identification of risky objects by masking the front agents. This formulation is also adopted by the STEEK model \cite{DBLP:journals/corr/abs-2111-09094} for the intention decision model (\emph{e.g.}, \emph{Stopping} or \emph{Moving Forward}), where “\emph{region-targeted counterfactual explanations}” is introduced and could generate meaningful counterfactuals with a preserved scene layout and relevant traffic light changing. In addition, some recent works \cite{zhang2022adversarial,cao2022advdo} begin to investigate the robustness of future prediction by attacking the input observations. These approaches aim the explainability by changing the semantic or scene state and checking the influence on the outcome for finding the primary input state. Actually, causal relation has been observed by the safety-critical driving scenario generation, such as the CausalAF \cite{dingcausalaf} that aligns with the behavioral graphs. CausalAF integrates the Causal Order Masks (COM) to generate possible cause-effect relations for the road scene and the Causal Visible Mask (CVM) to filter the non-causal information. The causality has a natural relationship with the social interaction of road agents. Therefore, the causality does not just correlate with the static road entities, but also the dynamic action or pose of the agents.

\subsection{Promising BIP-aware Applications}
From the investigation, we find that there are few research efforts on BIP-aware behavior prediction. Behavioral intention is the most direct promotion for certain behaviors and can enlarge the Time-to-Collision (TTC) for collision avoidance. 

The BIP problem has a direct link with risk assessment in driving, as shown in Fig. \ref{fig2}. Recently, the collision risk prediction work \cite{DBLP:conf/ivs/SchoonbeekPAD22} is modeled by inferring the hidden intention of surrounding objects. Similarly, Kim \emph{et al.} \cite{kim2019crash} learn to identify dangerous vehicles using a simulator, which learns the crash patterns in the real accident video data and constructs a GTACrash dataset. The crash label is refined by predicting the future paths of other vehicles. VIENA$^2$ \cite{aliakbarian2018viena} is a promising benchmark with the synthetic data for the prediction of crashes, pedestrian intention (\emph{e.g.}, \emph{Crossing}, \emph{Walking}, \emph{Stopping}), and front car's intention (\emph{e.g.}, \emph{Stopping}, \emph{Turning Right/Left}, and \emph{Left/Right Lane Changing}). 

In addition, Vehicle-to-Vehicle (V2V) or Vehicle-to-Anying (V2X) cooperation (internet of vehicles) \cite{DBLP:journals/arobots/GaoGLZ22} and road-vehicle collaboration \cite{DBLP:journals/tim/WangYCCL22} are promising applications with the help of other vehicles' perception and large-scale cloud data. For instance, some attempts \cite{DBLP:journals/corr/abs-2009-10868,chen2022intention} predict the pedestrian crossing intention from the cooperative vehicles' view. This kind of formulation can capture a larger range for road structure representation than a single vehicle's view. Within these applications, consistent and shared behavioral intention understanding is an important problem. For example, the LC intention for a vehicle may be understood as a Vehicle Overtaking (VO) intention because of the location difference for the Ego Vehicle (EV). Therefore, group-wise consistent understanding \cite{li2022v2x} in collaboration is promising with a reasonable spatial and temporal perception window partitioning. DeepAccident \cite{DBLP:journals/corr/abs-2304-01168} is a new dataset for accident understanding in virtual V2V scenarios, where the collision and the trajectories of vehicles are annotated, which may be useful for BIP-aware crash anticipation in V2V situations. 

\section{Conclusion}
This paper presents a comprehensive review of Behavioral Intention Prediction (BIP), with an investigation of the datasets, intention types, key factors, challenges, agent-centric BIP, and BIP-aware prediction applications. With the definition of different prediction tasks and the introduction of available datasets, the key factors and challenges are summarized from the aspects of road structure representation, social interaction modeling, and prediction uncertainty. Based on this, we chronologically review the pedestrian- and vehicle-centric BIPs from different modeling approaches. The BIP-aware trajectory prediction and behavior prediction are described and the BIP-aware behavior prediction has a large space to be developed. With a one-to-one response to potential challenges and possible insights, we discuss the theories and benchmarks, counterfactual analysis, parallel testing, and promising BIP-aware applications. We hope this survey can provide a good promotion for future BIP research.

\label{fcon1}
{\small
\bibliographystyle{IEEEtran}
\bibliography{ref}
}

\end{document}